\documentclass{article}
\pdfoutput=1
\usepackage{times}  
\usepackage{helvet} 
\usepackage{courier}  
\usepackage[hyphens]{url}  
\usepackage{graphicx} 
\usepackage{graphicx}  
 \newlength\titlebox 
\usepackage{natbib}
\usepackage{amsmath}
\usepackage{amsthm}
\usepackage{amsfonts}
\usepackage{algorithm2e}
\usepackage{booktabs}

\newcommand{\E}{\mathbb{E}} 
 
\newcommand{\Var}{\mathrm{Var}}

\newcommand{\refeqAT}{(7)}
\newcommand{\refeqbT}{(8)}

\newcommand{\refeqBoyanAT}{(17)}
\newcommand{\refeqpropbias}{(18)}

\newcommand{\refthrmrecursive}{{1}}
\newcommand{\reflemmahypothetical}{{1}}
\newcommand{\refthrmconvergence}{{2}}
\newcommand{\reflemmaBoyan}{{2}}

\newcommand{\refpropbiasvar}{{1}}

\newcommand{\refsecexp}{{4}}

\newcommand{\bspi}{{\boldsymbol{\pi}}}
\newcommand{\bsphi}{{\boldsymbol{\phi}}}
\newcommand{\bsPhi}{{\boldsymbol{\Phi}}}

\newcommand{\bstheta}{{\boldsymbol{\theta}}}

\newcommand{\bfb}{{\mathbf{b}}}

\newcommand{\bfr}{{\mathbf{r}}}

\newcommand{\bfz}{{\mathbf{z}}}

\newcommand{\bfA}{{\mathbf{A}}}

\newcommand{\bfI}{{\mathbf{I}}}

\newcommand{\bfP}{{\mathbf{P}}}

\title{Supplementary material for \protect\textit{Uncorrected least-squares temporal difference with lambda-return}}

\author{Takayuki Osogami\\
IBM Research - Tokyo\\
{\tt osogami@jp.ibm.com}}

\begin{document}




\newpage
\setcounter{page}{1}
\setcounter{equation}{19}
\setcounter{figure}{1}

\maketitle

\begin{abstract}
Here, we provide a supplementary material for Takayuki Osogami,
``Uncorrected least-squares temporal difference with lambda-return,''
which appears in {\it Proceedings of the 34th AAAI Conference on
Artificial Intelligence} (AAAI-20)
\citep{LSTD}.
\end{abstract}

\appendix

\section{Proofs}
\label{sec:proof}

In this section, we prove Theorem~\refthrmrecursive,
Lemma~\reflemmahypothetical, Theorem~\refthrmconvergence,
Lemma~\reflemmaBoyan, and Proposition~\refpropbiasvar.  Note
that equations (1)-(19) refers to those in \cite{LSTD}.

\subsection{Proof of Theorem~\refthrmrecursive}

From \refeqAT-\refeqbT, we have the following equality:
\begin{align}
  \bfA^{\rm Unc}_{T+1}
  & = \sum_{t=0}^{T} \bsphi_t \left(
  \bsphi_t -
  (1-\lambda)\,\gamma\sum_{m=1}^{T-t} (\lambda\,\gamma)^{m-1} \, \bsphi_{t+m}
  \right)^\top \\
  & = \sum_{t=0}^{T-1} \bsphi_t \left(
  \bsphi_t -
  (1-\lambda)\,\gamma\sum_{m=1}^{T-t} (\lambda\,\gamma)^{m-1} \, \bsphi_{t+m}
  \right)^\top 
  + \bsphi_T \, \bsphi_T^\top \\
  & = \sum_{t=0}^{T-1} \bsphi_t \Bigg(
  \bsphi_t
  - (1-\lambda) \, \gamma\sum_{m=1}^{T-t-1} (\lambda\,\gamma)^{m-1} \, \bsphi_{t+m}
  - (1-\lambda) \, \gamma (\lambda\,\gamma)^{T-t-1} \, \bsphi_T
  \Bigg)^\top + \bsphi_T \, \bsphi_T^\top \\
  & = \bfA^{\rm Unc}_T
  - \sum_{t=0}^{T-1} \bsphi_t (1-\lambda) \, \gamma \, (\lambda\,\gamma)^{T-t-1} \, \bsphi_{T}^\top
  + \bsphi_T \, \bsphi_T^\top \\
  & = \bfA^{\rm Unc}_T
  + \left(\sum_{t=0}^T(\lambda\,\gamma)^{T-t}\,\bsphi_t\right)\bsphi_T^\top 
  - \gamma \left(\sum_{t=0}^{T-1} (\lambda\,\gamma)^{T-t-1}\,\bsphi_t\right)\bsphi_T^\top \label{eq:beforeAT}\\
  & = \bfA^{\rm Unc}_T + (\bfz_T - \gamma \, \bfz_{T-1}) \, \bsphi_T^\top. \label{eq:afterAT}\\
  \bfb_{T+1}
  & = \sum_{t=0}^T \bsphi_t \sum_{m=0}^{T-t} (\lambda\,\gamma)^m \, r_{t+1+m}\\
  & = \sum_{t=0}^{T-1} \bsphi_t
  \left( \sum_{m=0}^{T-t-1} (\lambda\,\gamma)^m \, r_{t+1+m} + (\lambda\,\gamma)^{T-t} \, r_{T+1} \right)
  + \bsphi_T \, r_{T+1} \label{eq:beforebT}\\
  & = \bfb_T + \bfz_T \, r_{T+1}. \label{eq:afterbT}
\end{align}
Here, the equality from \eqref{eq:beforeAT} to \eqref{eq:afterAT} and
the equality from \eqref{eq:beforebT} to \eqref{eq:afterbT} follow
from the definition of the eligibility trace $\bfz_T$ in the theorem.
The recursive computation of the eligibility trace can be verified in
a straightforward manner.  This completes the proof of
Theorem~\refthrmrecursive.

\subsection{Proof of Lemma~\reflemmahypothetical}

Observe that there exists $T_0$ such that $\frac{1}{T}\bfA_T^{\rm
  Unc}$ is invertible for any $T>T_0$, because we assume that
$\frac{1}{T}\bfA_T^{\rm Unc}$ converges to an invertible matrix as
$T\to\infty$, and invertible matrices form an open set.
Then for each $T>T_0$, we have
\begin{align}
  \frac{1}{T} \bfA_T^{\rm Unc} \, (\bstheta_T - \bstheta_T^\star)
  & = \frac{1}{T} (\bfb_T - \bfb_T^\star) \\
    (\bstheta_T - \bstheta_T^\star)
  & = \left(\frac{1}{T} \bfA_T^{\rm Unc}\right)^{-1}
  \, \frac{1}{T} (\bfb_T - \bfb_T^\star).
\end{align}
By the continuity of matrix inverse, we then have
\begin{align}
  \lim_{T\to\infty} (\bstheta_T - \bstheta_T^\star)
  & = \left(\lim_{T\to\infty} \frac{1}{T} \bfA_T^{\rm Unc}\right)^{-1}
  \lim_{T\to\infty} \frac{1}{T} (\bfb_T - \bfb_T^\star).
\end{align}
It thus suffices to show $\frac{1}{T}|\bfb_T^\star-\bfb_T| \to 0$ as
$T\to\infty$.

Because the state space ${\cal S}$ is finite, the magnitude of the
immediate reward and the feature vector is uniformly bounded.  Namely,
there exists $c<\infty$ such that $R(s)\le c$ and $|\bsphi(s)| \le c$
elementwise for any $s\in{\cal S}$.  Thus, we have the following
elementwise inequality:
\begin{align}
  \frac{1}{T}|\bfb_T^\star-\bfb_T|
  & = \frac{1}{T} \left|\sum_{t=0}^{T-1} \bsphi_t \, \lambda^{T-t-1} \sum_{m=T-t}^\infty \gamma^m \sum_{s'\in{\cal S}} (\bfP^m)_{s,s'} \, R(s') \right| \\
  & \le c^2 \, \frac{1}{T} \sum_{t=0}^{T-1} \lambda^{T-t-1} \sum_{m=T-t}^\infty \gamma^m \sum_{s'\in{\cal S}} (\bfP^m)_{s,s'} \\  
  & = c^2 \, \frac{1}{T} \sum_{t=0}^{T-1} \lambda^{T-t-1} \sum_{m=T-t}^\infty \gamma^m \\  
  & = c^2 \, \frac{1}{T} \sum_{t=0}^{T-1} \lambda^{T-t-1} \, \frac{\gamma^{T-t}}{1-\gamma} \\
  & = c^2 \, \frac{\gamma}{1-\gamma} \, \frac{1}{T} \sum_{t=0}^{T-1} (\lambda\,\gamma)^{T-t-1} \\
  & = c^2 \, \frac{\gamma}{1-\gamma} \, \frac{1}{T} \, \frac{1-(\lambda\,\gamma)^T}{1-\lambda\,\gamma},
\end{align}
which tends to 0 as $T\to\infty$.  This completes the proof of
Lemma~\reflemmahypothetical.

\subsection{Proof of Theorem~\refthrmconvergence}

At each step $T$, Uncorrected LSTD($\lambda$) gives the weights $\bstheta_T$,
which is the solution of $\frac{1}{T}\bfA_T^{\rm Unc} \, \bstheta = \frac{1}{T}\bfb_T$.
Therefore, it suffices to show $\frac{1}{T}\bfA_T^{\rm Unc}\to\bar\bfA$
and $\frac{1}{T}\bfb_T\to\bar\bfb$ as $T\to 0$.

Due to the ergodicity of the Markov chain, as $T\to\infty$, each state
is visited infinitely often, and the time each state is occupied is
proportional to the steady state probability almost surely.  Then, by
the pointwise ergodic theorem, we have the following almost sure
convergence:
\begin{align}
  \lim_{T\to\infty} \frac{1}{T} \bfA^{\rm Unc}_T
  & = \lim_{T\to\infty} \frac{1}{T} \sum_{t=0}^{T-1} \bsphi_t \left(
  \bsphi_t -
  (1-\lambda)\,\gamma\sum_{m=1}^{T-t-1} (\lambda\,\gamma)^{m-1} \, \bsphi_{t+m}
  \right)^\top \label{eq:timeAt}\\
  & = \sum_{s\in{\cal S}} \bspi(s) \, \bsphi(s) \Bigg(
  \sum_{m=0}^\infty (\lambda\,\gamma)^m \sum_{s'\in{\cal S}} (\bfP^m)_{s,s'} \, \bsphi(s')
  - \gamma\sum_{m=1}^\infty (\lambda\,\gamma)^{m-1} \sum_{s'\in{\cal S}} (\bfP^m)_{s,s'} \, \bsphi(s')
  \Bigg)^\top \label{eq:ensembleAt}\\
  & = \bsPhi^\top {\rm Diag}(\bspi) \, (\bfI-\lambda\,\gamma\,\bfP)^{-1} \, \bsPhi
  - \gamma \, \bsPhi^\top {\rm Diag}(\bspi) \, \bfP \, (\bfI-\lambda\,\gamma\,\bfP)^{-1} \, \bsPhi \\
  & = \bsPhi^\top {\rm Diag}(\bspi) \, (\bfI-\gamma\,\bfP) \, (\bfI-\lambda\,\gamma\,\bfP)^{-1} \, \bsPhi
\end{align}
and
\begin{align}
  \lim_{T\to\infty} \frac{1}{T} \bfb_T
  & = \lim_{T\to\infty} \frac{1}{T} \sum_{t=0}^{T-1} \bsphi_t \sum_{m=0}^{T-t-1} (\lambda\,\gamma)^m \, r_{t+1+m}
  \label{eq:timebt}\\
  & = \sum_{s\in{\cal S}} \bspi(s) \, \bsphi(s) \sum_{m=0}^\infty (\lambda\,\gamma)^m \sum_{s'\in{\cal S}} (\mathbf{P}^m)_{s,s'} \, R(s')
  \label{eq:ensemblebt}\\
  & = \sum_{m=0}^\infty (\lambda\,\gamma)^m \, \bsPhi^\top {\rm Diag}(\bspi) \, \bfP^m \, \bfr\\
  & = \bsPhi^\top {\rm Diag}(\bspi) \, (\bfI - \lambda\,\gamma\,\bfP)^{-1} \, \bfr,
\end{align}
which establishes the theorem.  Here, the equality from
\eqref{eq:timeAt} to \eqref{eq:ensembleAt} and the equality from
\eqref{eq:timebt} to \eqref{eq:ensemblebt} relate the time average to
the ensemble average (almost surely) via the pointwise ergodic
theorem.

\subsection{Proof of Lemma~\reflemmaBoyan}

From \refeqBoyanAT, we have
\begin{align}
  \bfA^{\rm Boy}_T
  & = \sum_{t=0}^{T-1} \bsphi_t \Bigg(
  \bsphi_t -
  (1-\lambda)\,\gamma\sum_{m=1}^{T-t-1} (\lambda\,\gamma)^{m-1} \, \bsphi_{t+m}
  - \gamma \, (\lambda\,\gamma)^{T-t-1} \, \bsphi_{T}
  \Bigg)^\top \\
  & = \sum_{t=0}^{T-1} \bsphi_t \left(
  \sum_{m=0}^{T-t-1} (\lambda\,\gamma)^m \, \bsphi_{t+m}
  - \gamma\sum_{m=1}^{T-t} (\lambda\,\gamma)^{m-1} \, \bsphi_{t+m}
  \right)^\top \\
  & = \sum_{t=0}^{T-1} \bsphi_t \left(
  \sum_{k=t}^{T-1} (\lambda\,\gamma)^{k-t} \, \bsphi_k
  - \gamma\sum_{k=t+1}^{T} (\lambda\,\gamma)^{k-t-1} \, \bsphi_k
  \right)^\top \\
  & =
  \sum_{k=0}^{T-1} \sum_{t=0}^k
  (\lambda\,\gamma)^{k-t} \, \bsphi_t \, \bsphi_k^\top
  - \gamma \sum_{k=1}^T \sum_{t=0}^{k-1}
  (\lambda\,\gamma)^{k-t-1} \, \bsphi_t \, \bsphi_k^\top \label{eq:beforeBoyan}\\
  & = \sum_{k=0}^{T-1} \bfz_k \, \bsphi_k^\top
  - \gamma \sum_{k=1}^T \bfz_{k-1} \, \bsphi_k^\top \label{eq:afterBoyan}\\
  & = \sum_{k=0}^{T-1} \bfz_k \, (\bsphi_k - \gamma \, \bsphi_{k+1})^\top,
\end{align}
where the equality from \eqref{eq:beforeBoyan} to
\eqref{eq:afterBoyan} follows from the definition of the eligibility
trace in Theorem~\refthrmrecursive.  This completes the proof of
the lemma.

\subsection{Proof of Proposition~\refpropbiasvar}

  Because there is no transition of states, we can let $\bsphi_t=1,
  \forall t$.  Then the coefficient matrix \refeqBoyanAT for
  Boyan's LSTD($\lambda$) is reduced to the following one dimensional
  constant for given $\lambda$, $\gamma$, and $T$:
  \begin{align}
    A_T^{\rm Boy}
    & = \sum_{t=0}^{T-1}
    \left(
    1 - (1-\lambda) \, \gamma \sum_{m=1}^{T-t-1} (\lambda\,\gamma)^{m-1} - \gamma\,(\lambda\gamma)^{T-t-1}
    \right) \\
    & = (1-\gamma) \sum_{t=0}^{T-1} \sum_{m=0}^{T-t-1} (\lambda\gamma)^m \\
    & = (1-\gamma) \sum_{n=1}^T \frac{1-(\lambda\gamma)^n}{1-\lambda\gamma}.
    \label{eq:prop:BoyanAT}\\
    & = \frac{(1-\gamma)\,T}{1-\lambda\gamma} + o(T).
  \end{align}
  From \refeqAT and \refeqBoyanAT, we have
  \begin{align}
    A_T^{\rm Unc}
    & = \sum_{t=0}^{T-1}
    \left(
    1 - (1-\lambda) \, \gamma \sum_{m=1}^{T-t-1} (\lambda\,\gamma)^{m-1}
    \right) \\
    & = A_T^{\rm Boy}  + \gamma \sum_{t=0}^{T-1} (\lambda\gamma)^{T-t-1} \\
    & = A_T^{\rm Boy}  + \gamma \, \frac{1-(\lambda\gamma)^T}{1-\lambda\gamma}.
    \label{eq:prop:AT}
  \end{align}
  Let
  \begin{align}
    \Delta_T
    & \equiv A_T^{\rm Unc} - A_T^{\rm Boy}\\
    & = \gamma \, \frac{1-(\lambda\gamma)^T}{1-\lambda\gamma} \\
    & = \frac{\gamma}{1-\lambda\gamma} + o(1).
  \end{align}

  The estimator of the discounted cumulative reward is given by
  $\theta^{\rm Unc}_T=b_T / A_T^{\rm Unc}$
  or $\theta^{\rm Boy}_T=b_T / A_T^{\rm Boy}$,
  where $b_T$ is reduced to the following random variable
  (here, $R_{t+1+m}$ denotes the reward obtained at step $t+1+m$):
  \begin{align}
    b_T
    & = \sum_{t=0}^{T-1} \sum_{m=0}^{T-t-1} (\lambda\,\gamma)^m \, R_{t+1+m}\\
    & = \sum_{n=1}^T \sum_{t=0}^{n-1} (\lambda\,\gamma)^{n-t-1} \, R_n \\
    & = \sum_{n=1}^T \frac{1-(\lambda\,\gamma)^n}{1-\lambda\,\gamma} R_n.
  \end{align}
  where the second equality follows by changing variables $n=m+t+1$
  and exchanging the order of summations.  Because the reward is i.i.d., the expectation and
  variance of $b_T$ is given as follows:
  \begin{align}
    \E[b_T]
    & = \mu \sum_{n=1}^T \frac{1-(\lambda\,\gamma)^n}{1-\lambda\,\gamma}
    \label{eq:prop:EbT}\\
    \Var[b_T]
    & = \sigma^2 \sum_{n=1}^T \left(\frac{1-(\lambda\,\gamma)^n}{1-\lambda\,\gamma}\right)^2.
  \end{align}

  By \eqref{eq:prop:BoyanAT} and \eqref{eq:prop:EbT}, it is
  straightforward to verify that $\theta_T^{\rm Boy}$ is unbiased.
  Indeed, the expected value is
  \begin{align}
    \E[\theta_T^{\rm Boy}]
    & = \frac{\E[b_T]}{A_T^{\rm Boy}}
    = \frac{\mu}{1-\gamma},
    \label{eq:prop:unbiased}
  \end{align}
  which coincides with the true expected discount cumulative reward.

  Now, by \eqref{eq:prop:AT} and \eqref{eq:prop:unbiased}, the bias of $\theta_T^{\rm Unc}$ is given by
  \begin{align}
    \E[\theta_T^{\rm Unc}] - \frac{\mu}{1-\gamma}
    & = \frac{\E[b_T]}{A_T^{\rm Unc}} - \frac{\E[b_T]}{A_T^{\rm Boy}} \\
    & = \frac{A_T^{\rm Boy} - A_T^{\rm Unc}}{A_T^{\rm Unc}} \, \frac{\E[b_T]}{A_T^{\rm Boy}} \\
    & = -\frac{\Delta}{A_T^{\rm Boy} + \Delta} \, \frac{\mu}{1-\gamma} \\
    & = - \frac{\gamma\mu}{(1-\gamma)^2T} + o\left(\frac{1}{T}\right),
  \end{align}
  which establishes \refeqpropbias.

  Finally, the variance of the estimator is given by
  $\Var[\theta_T^{\rm Boy}] = \frac{\Var[b_T]}{(A_T^{\rm Boy})^2}$ and
  $\Var[\theta_T^{\rm Unc}] = \frac{\Var[b_T]}{(A_T^{\rm Unc})^2}$.  Hence,
  we have
  \begin{align}
    \frac{\Var[\theta_T^{\rm Boy}]}{\Var[\theta_T^{\rm Unc}]}
    & = \left(\frac{A_T^{\rm Boy}+\Delta_T}{A_T^{\rm Boy}}\right)^2 \\
    & = 1 + \frac{2\,\gamma}{(1-\gamma)\,T} + o\left(\frac{1}{T}\right).
  \end{align}
  This completes the proof of the proposition.

\section{Details of experiments}

In this section, we provide the details of the experiments in
Section~\refsecexp.

\subsection{Computational environment}

To generate the random MRPs and to run the experiments, we use the
library\footnote{https://github.com/armahmood/totd-rndmdp-experiments}
published by \citet{TrueTDb}.
We run our experiments on a Ubuntu 18.04 workstation having eight Intel
Core i7-6700K CPUs running at 4.00~GHz and 64~GB random access memory.

\subsection{Detailed results of experiments}

\begin{figure*}[tbh]
  \centering
  \underline{{\em small} MRP $(10, 3, 0.1)$}\\
  \begin{minipage}{0.32\linewidth}
    \includegraphics[width=\linewidth]{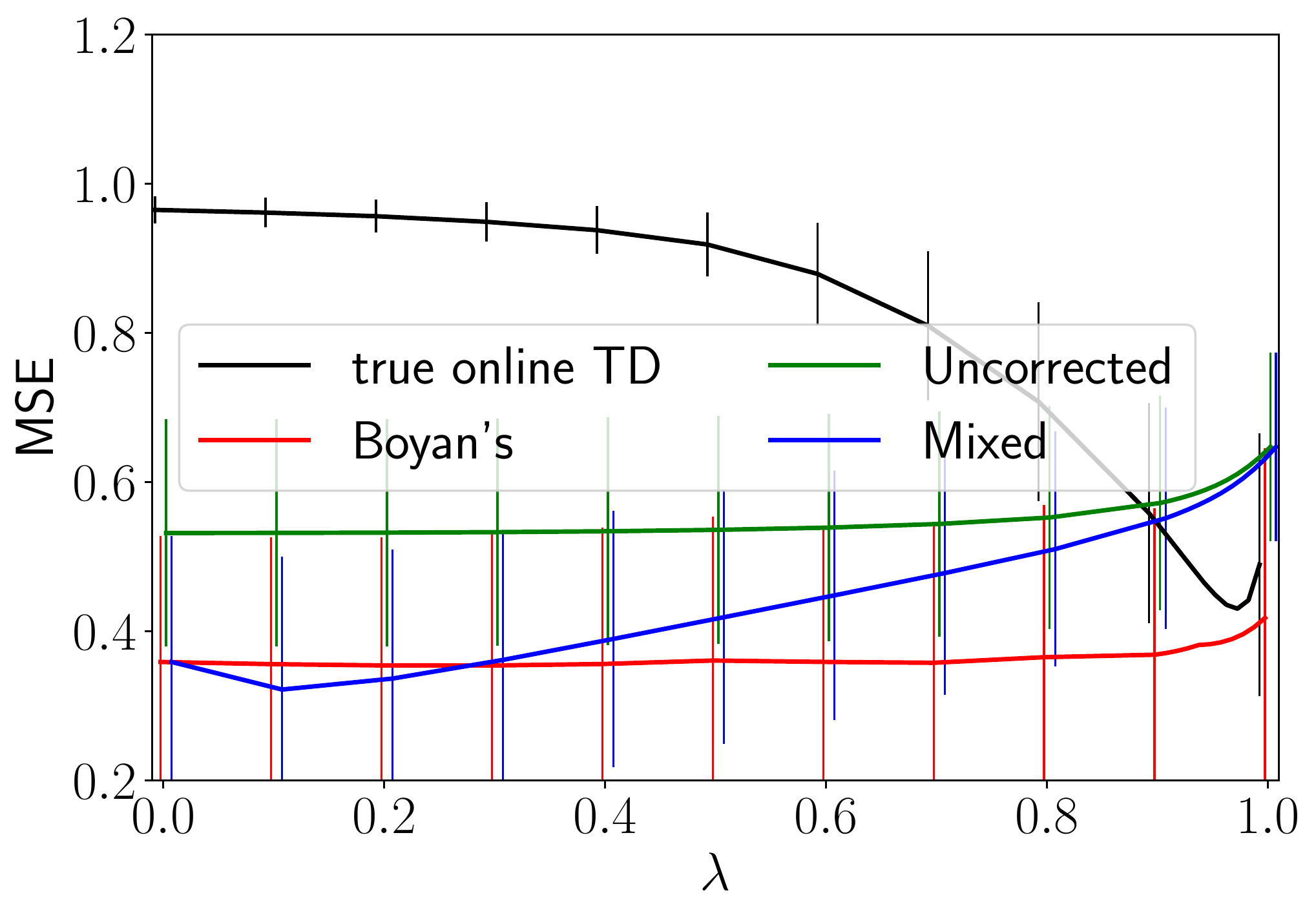}
  \end{minipage}
  \begin{minipage}{0.32\linewidth}
    \includegraphics[width=\linewidth]{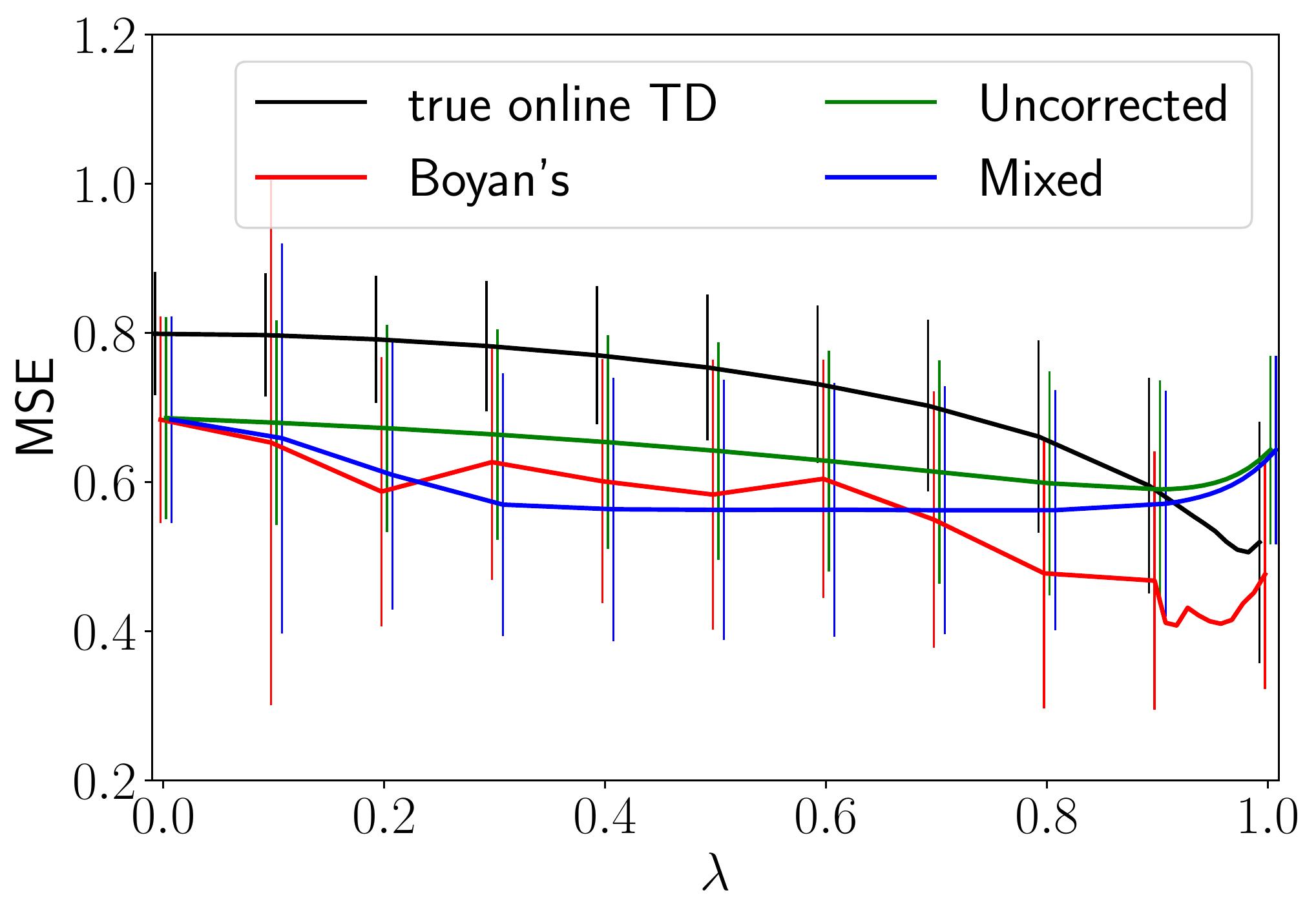}
  \end{minipage}
  \begin{minipage}{0.32\linewidth}
    \includegraphics[width=\linewidth]{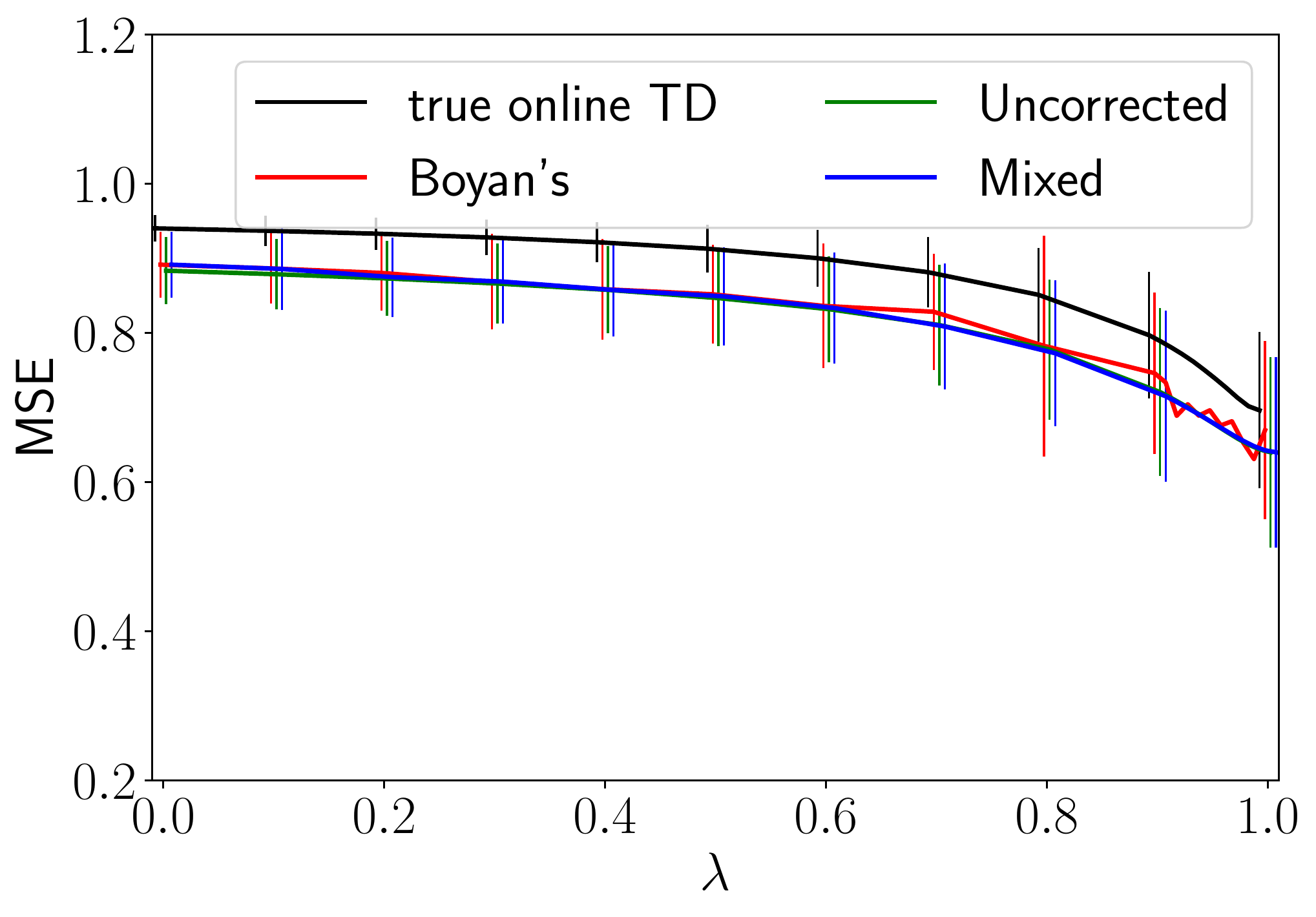}
  \end{minipage}
  \ \\
  \ \\
  \underline{{\em large} MRP $(100, 10, 0.1)$}\\
  \begin{minipage}{0.32\linewidth}
    \includegraphics[width=\linewidth]{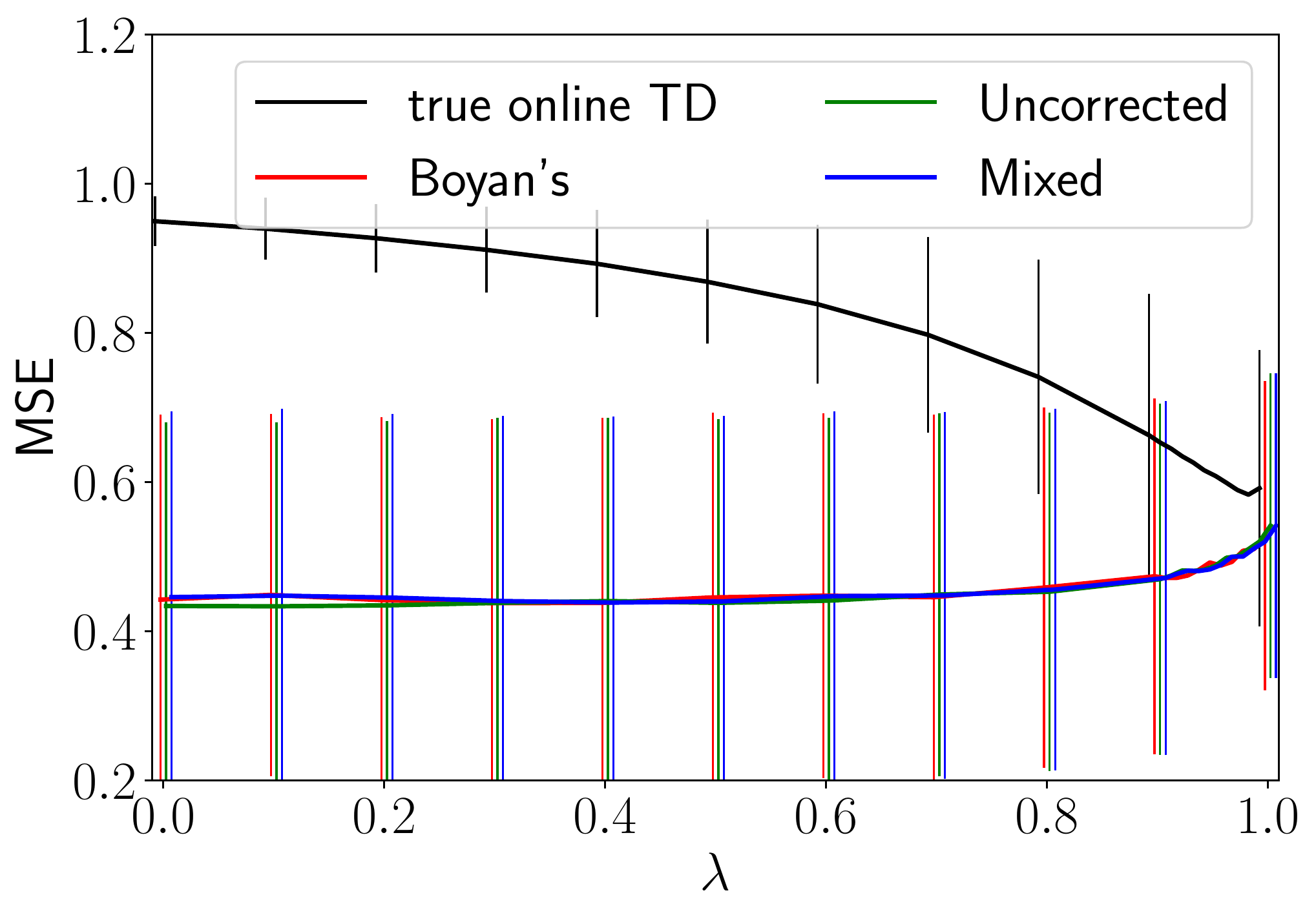}
  \end{minipage}
  \begin{minipage}{0.32\linewidth}
    \includegraphics[width=\linewidth]{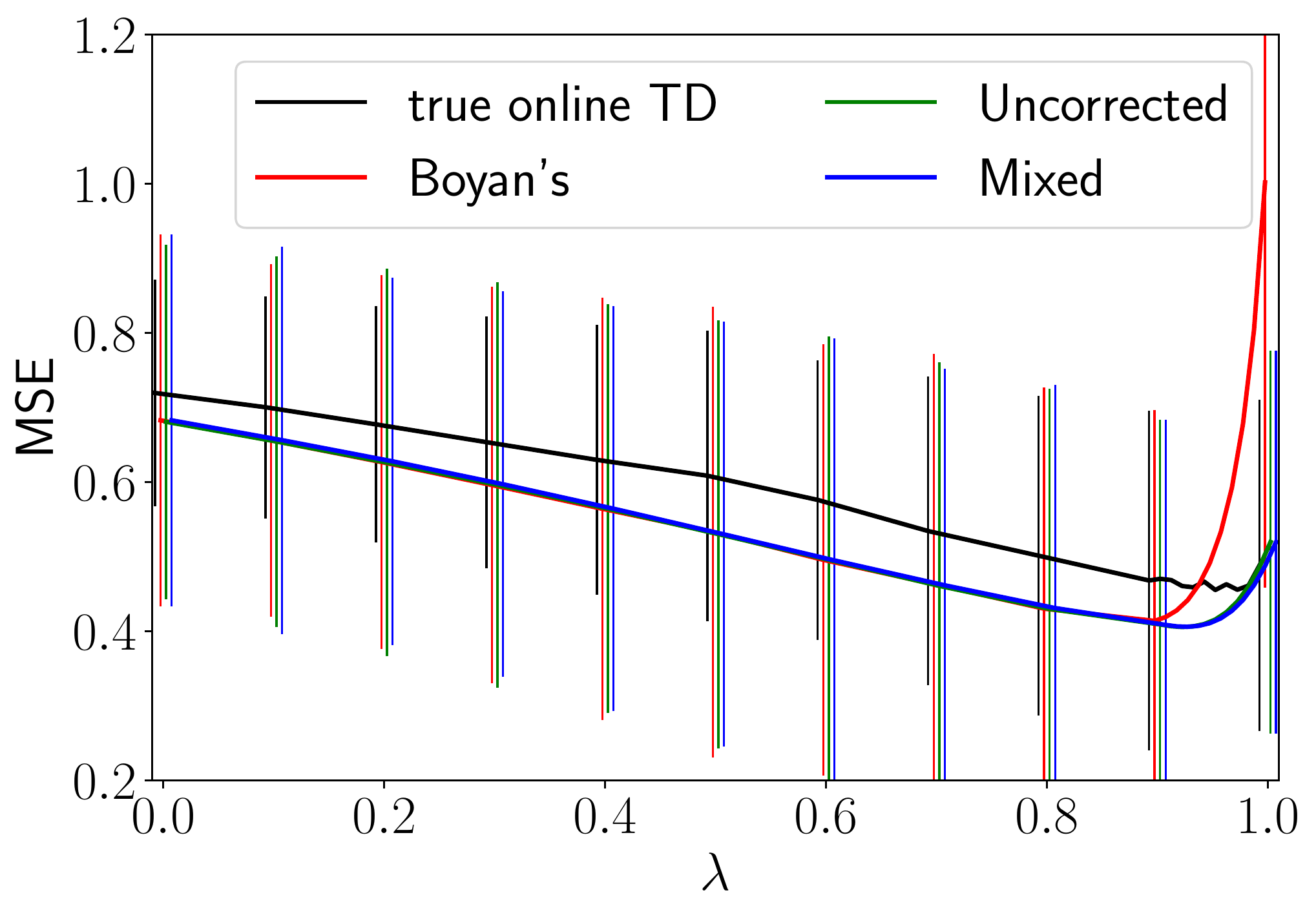}
  \end{minipage}
  \begin{minipage}{0.32\linewidth}
    \includegraphics[width=\linewidth]{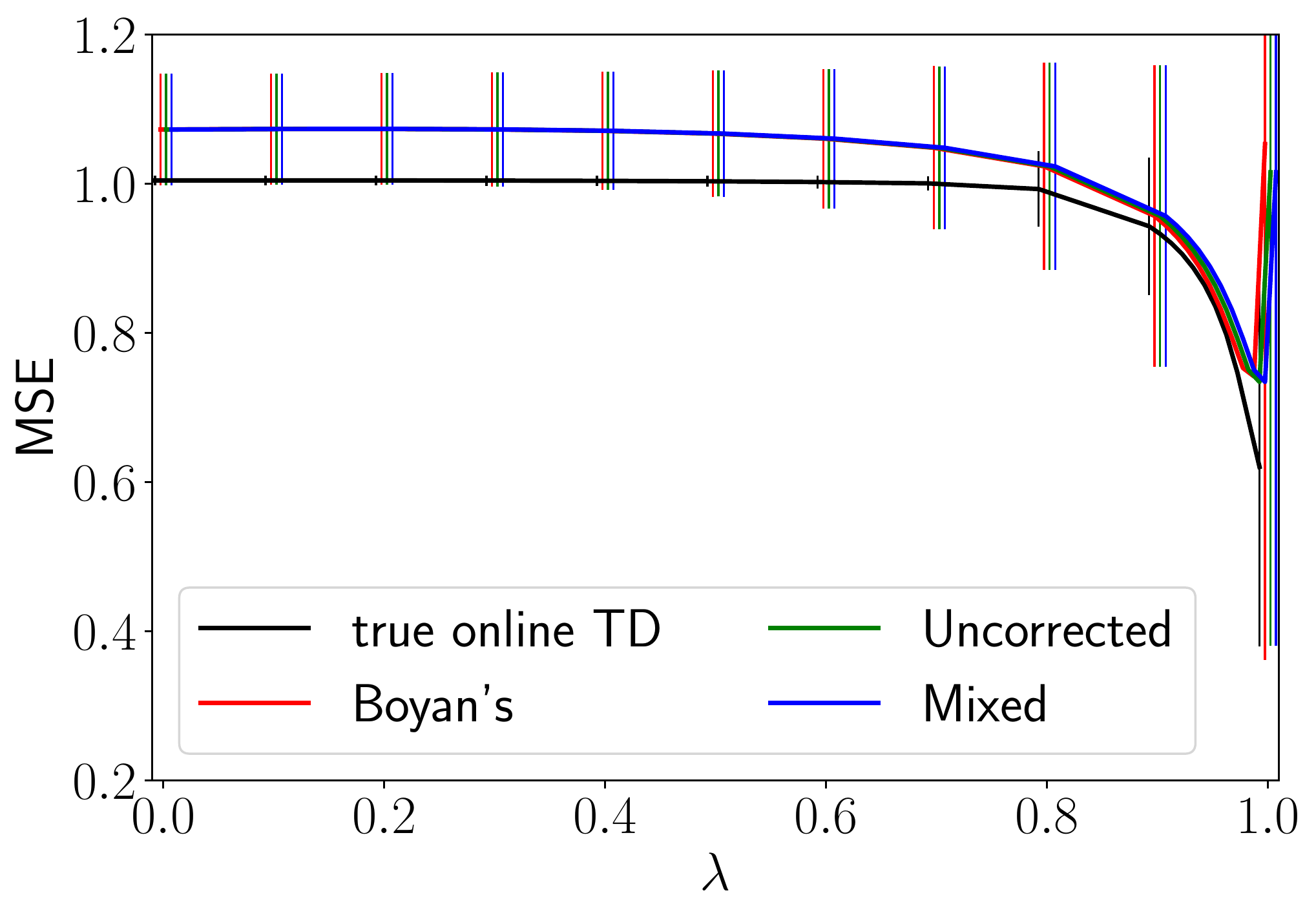}
  \end{minipage}
  \ \\
  \ \\
  \underline{{\em deterministic} MRP $(100, 3, 0)$}\\
  \begin{minipage}{0.32\linewidth}
    \centering
    \includegraphics[width=\linewidth]{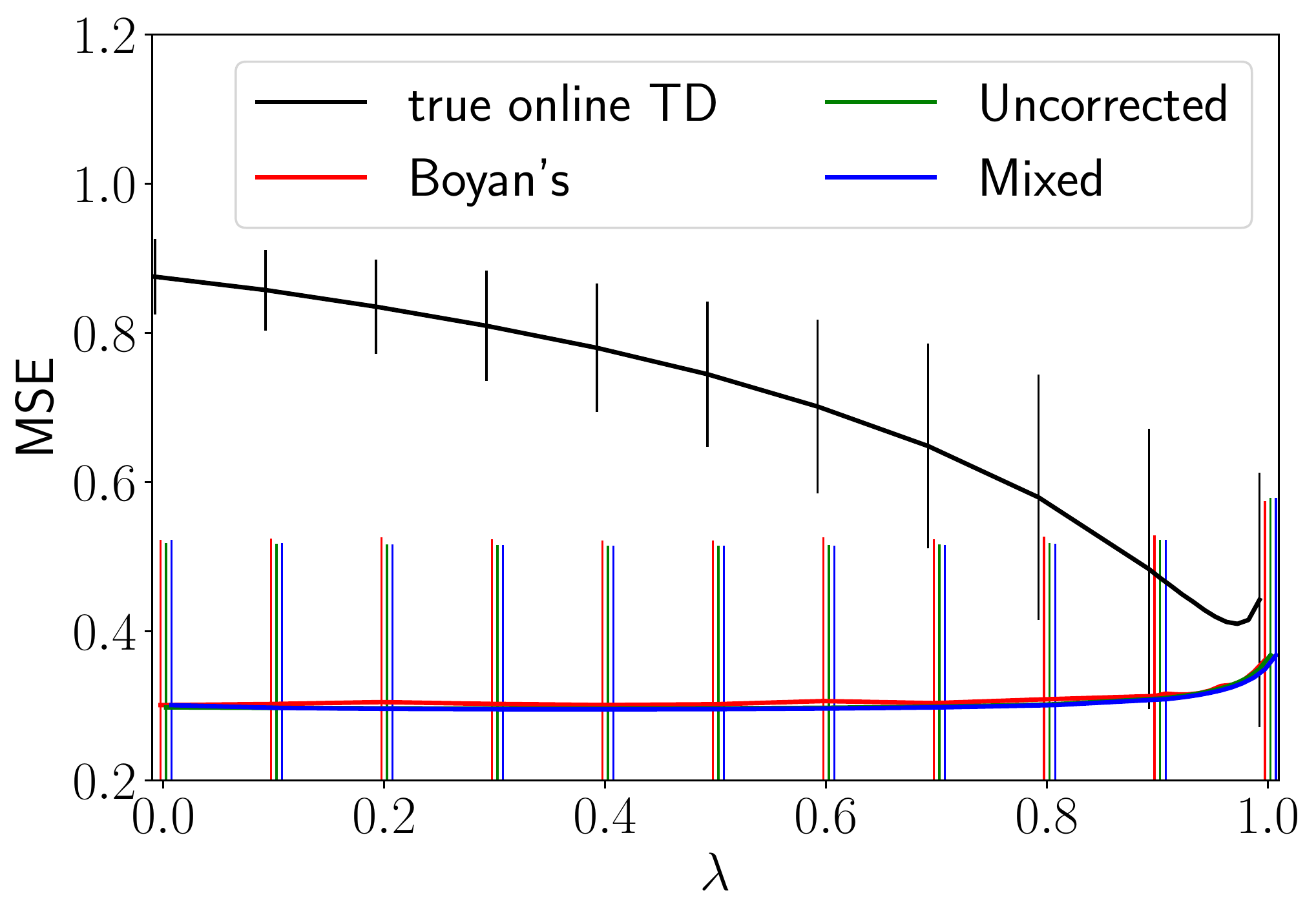}\\
    (a) {\em tabular} features
  \end{minipage}
  \begin{minipage}{0.32\linewidth}
    \centering
    \includegraphics[width=\linewidth]{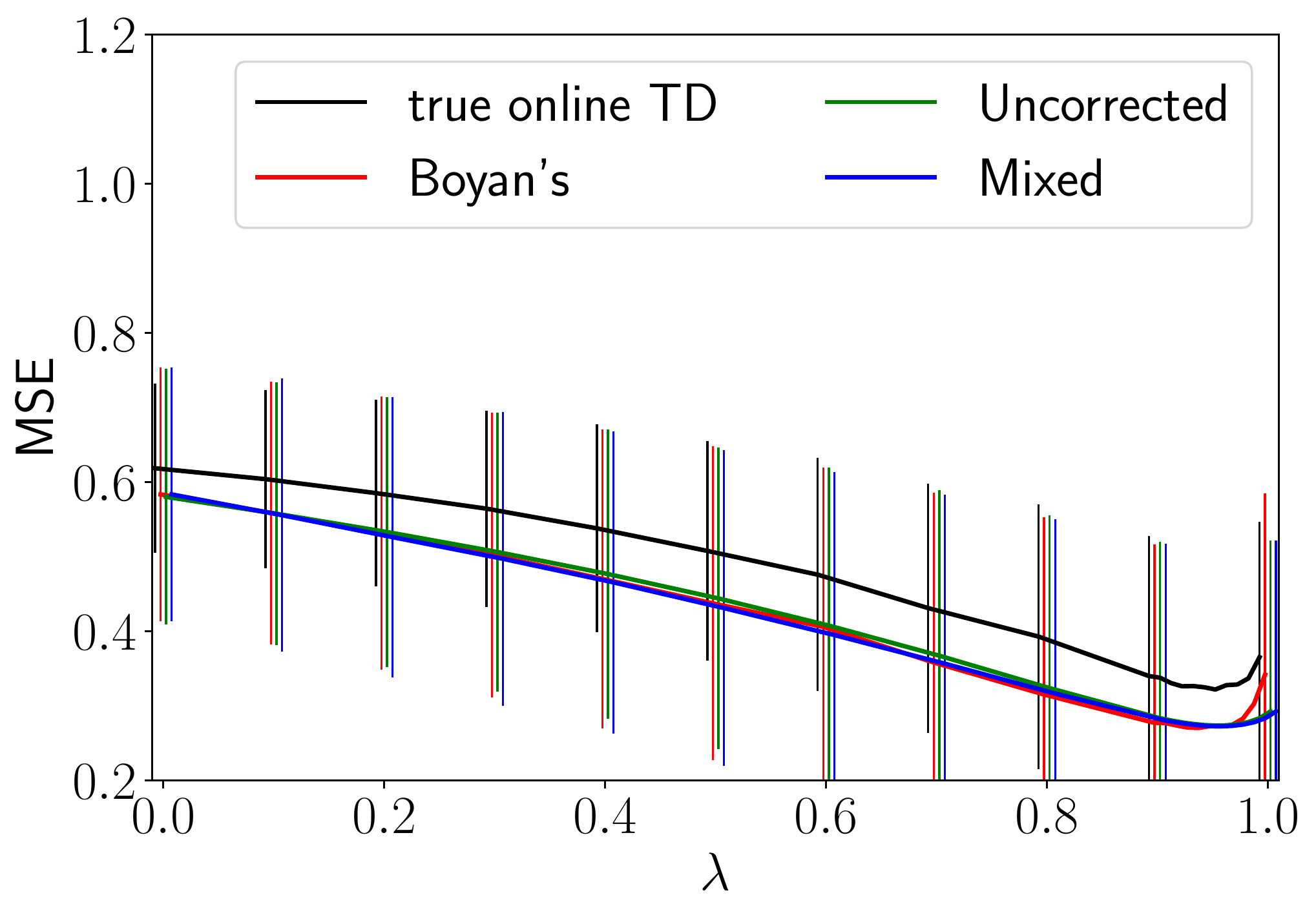}\\
    (b) {\em binary} features
  \end{minipage}
  \begin{minipage}{0.32\linewidth}
    \centering
    \includegraphics[width=\linewidth]{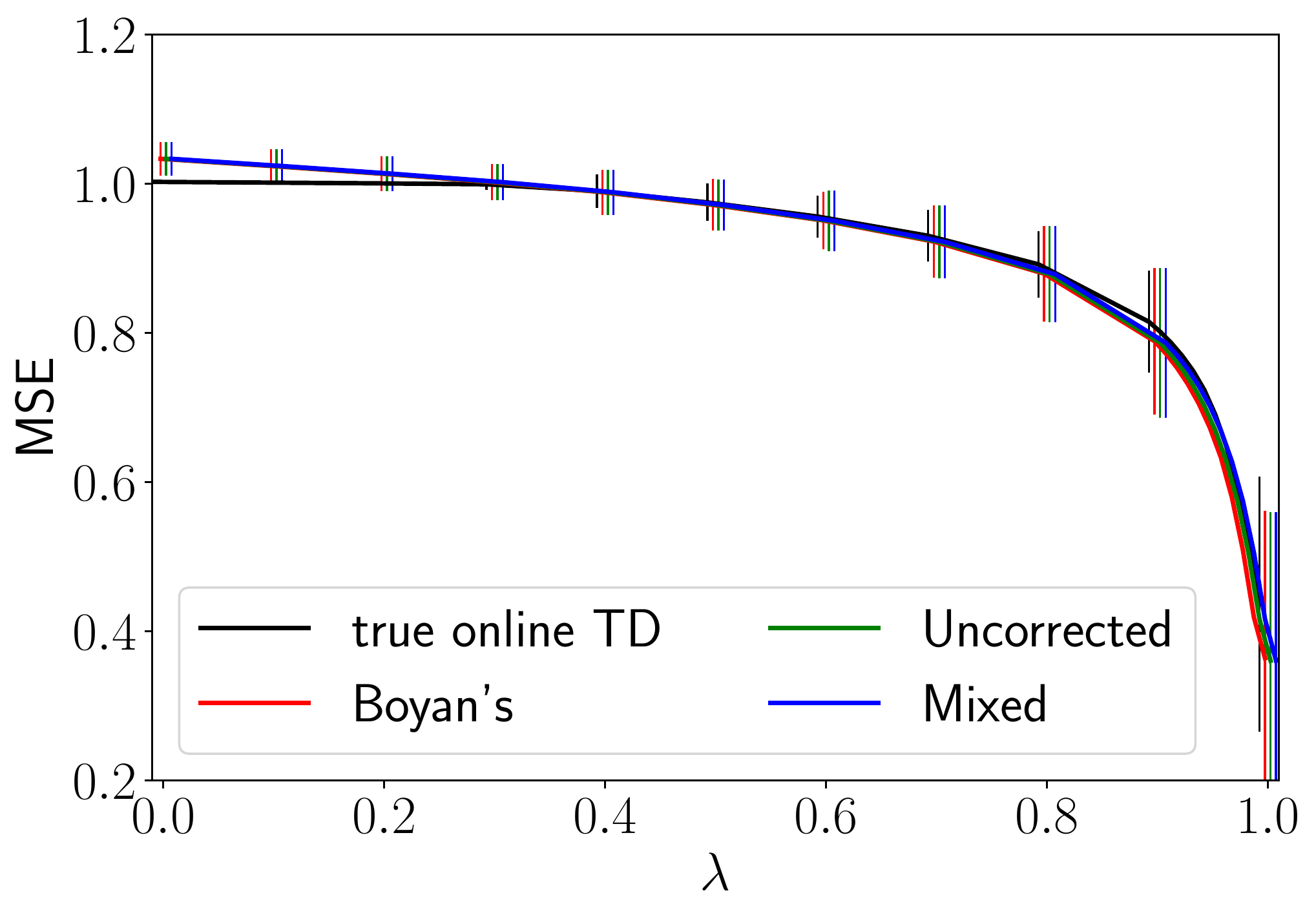}\\
    (c) {\em non-binary} features
  \end{minipage}
  \caption{Mean squared error (MSE) of Uncorrected, Mixed, and Boyan's
    LSTD($\lambda$) as well as true online TD($\lambda$).  The MSE is
    evaluated on three random Markov reward processes (MRPs), each
    with three representations (features) of states.  For Uncorrected,
    Mixed, and Boyan's LSTD($\lambda$), the best MSE with the optimal
    value of regularization coefficient (among $\{2^{i}\mid
    i=-8,-7,\ldots,7,8\}$) is shown for each $\lambda$.  For true
    online TD($\lambda$), the best MSE with the optimal step size (see
    \citet{TrueTDb} for details) is shown for each $\lambda$.  For
    each data point, MSE is computed on the basis of 50 runs, and the
    error bar shows its standard deviation.  For clarity, we show
    error bars only at $\lambda=i/10$ for $i=0, \ldots, 9, 10$.}
  \label{fig:errorbar}
\end{figure*}

Figure~\ref{fig:errorbar} shows the results corresponding to Figure~4
of \citet{TrueTDb}.  A difference is that, for the three
LSTD($\lambda$)s, we show the best MSE with the optimal value of
regularization coefficient, $\alpha$, among $\{2^{i}\mid
i=-8,-7,\ldots,7,8\}$ for each data point.  In Figure~4 of
\citet{TrueTDb}, the best MSE with the optimal step size is shown for
each TD($\lambda$).  As a reference we include the results with true
online TD($\lambda$) in Figure~\ref{fig:errorbar}.

Figure~\ref{fig:errorbar} shows the best achievable MSE with the
optimal choice of hyperparameters for each LSTD($\lambda$) and for
true online TD($\lambda$), but this best achievable MSE cannot be
achieved in practice, because one cannot optimally tune the
hyperparameters.

\begin{figure*}[hb]
  \paragraph{Detailed results with the {\em small} MRP}
  \ \\
  \ \\
  \begin{minipage}{0.2\linewidth}
    \underline{Mixed LSTD($\lambda$)}
  \end{minipage}
  \begin{minipage}{0.26\linewidth}
    \includegraphics[width=\linewidth]{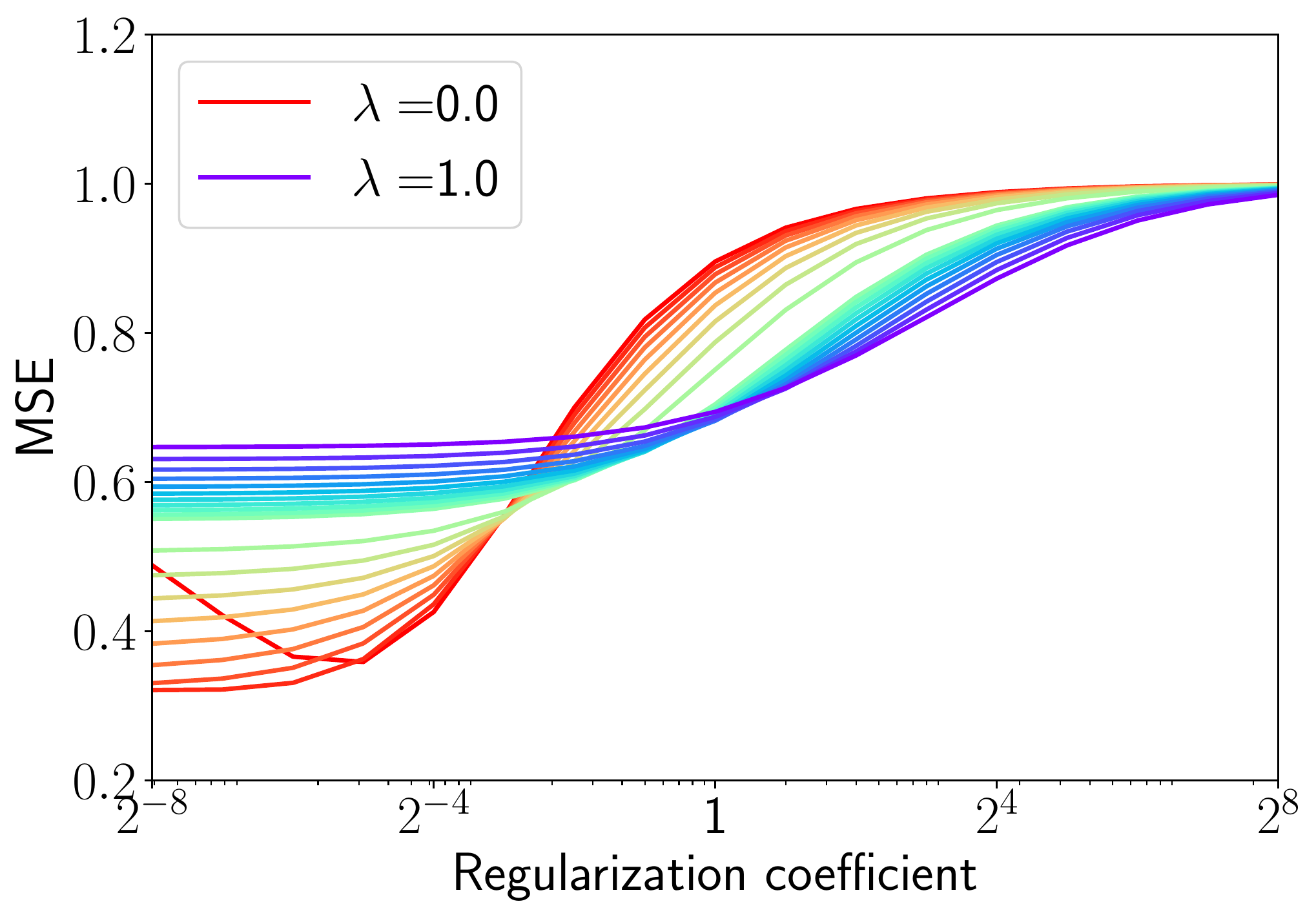}
  \end{minipage}
  \begin{minipage}{0.26\linewidth}
    \includegraphics[width=\linewidth]{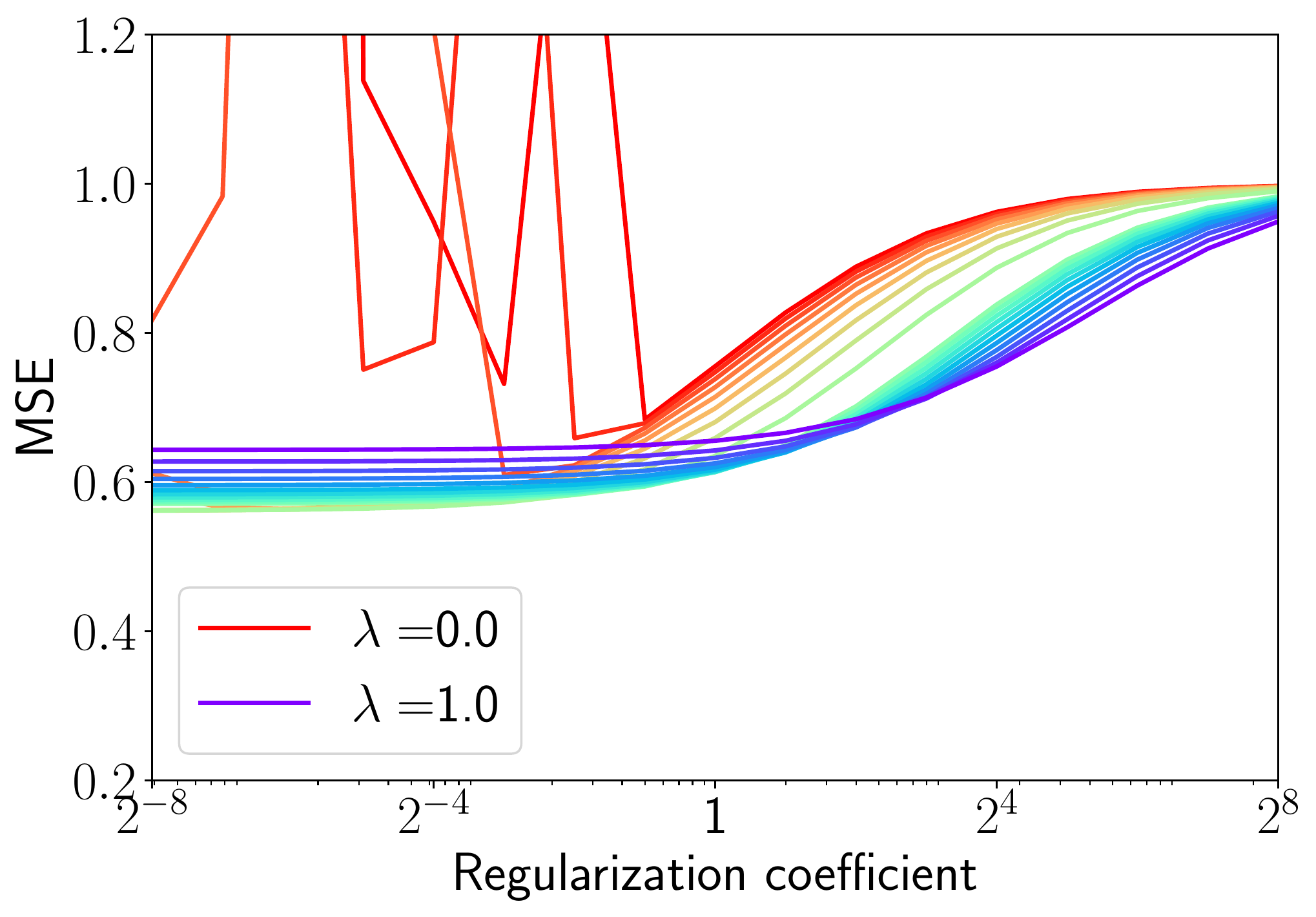}
  \end{minipage}
  \begin{minipage}{0.26\linewidth}
    \includegraphics[width=\linewidth]{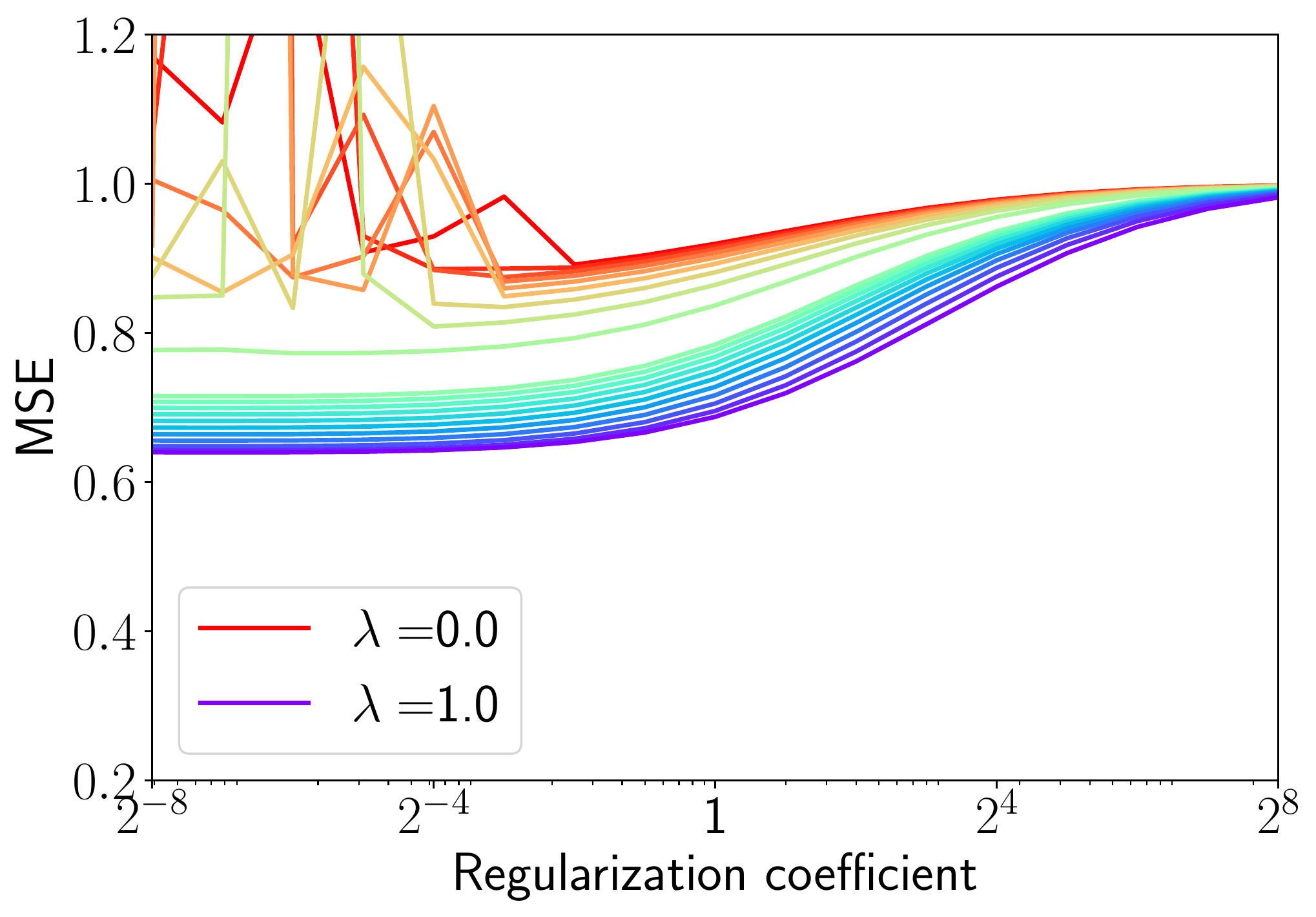}
  \end{minipage}
  \begin{minipage}{0.2\linewidth}
    \underline{Uncorrected LSTD($\lambda$)}
  \end{minipage}
  \begin{minipage}{0.26\linewidth}
    \includegraphics[width=\linewidth]{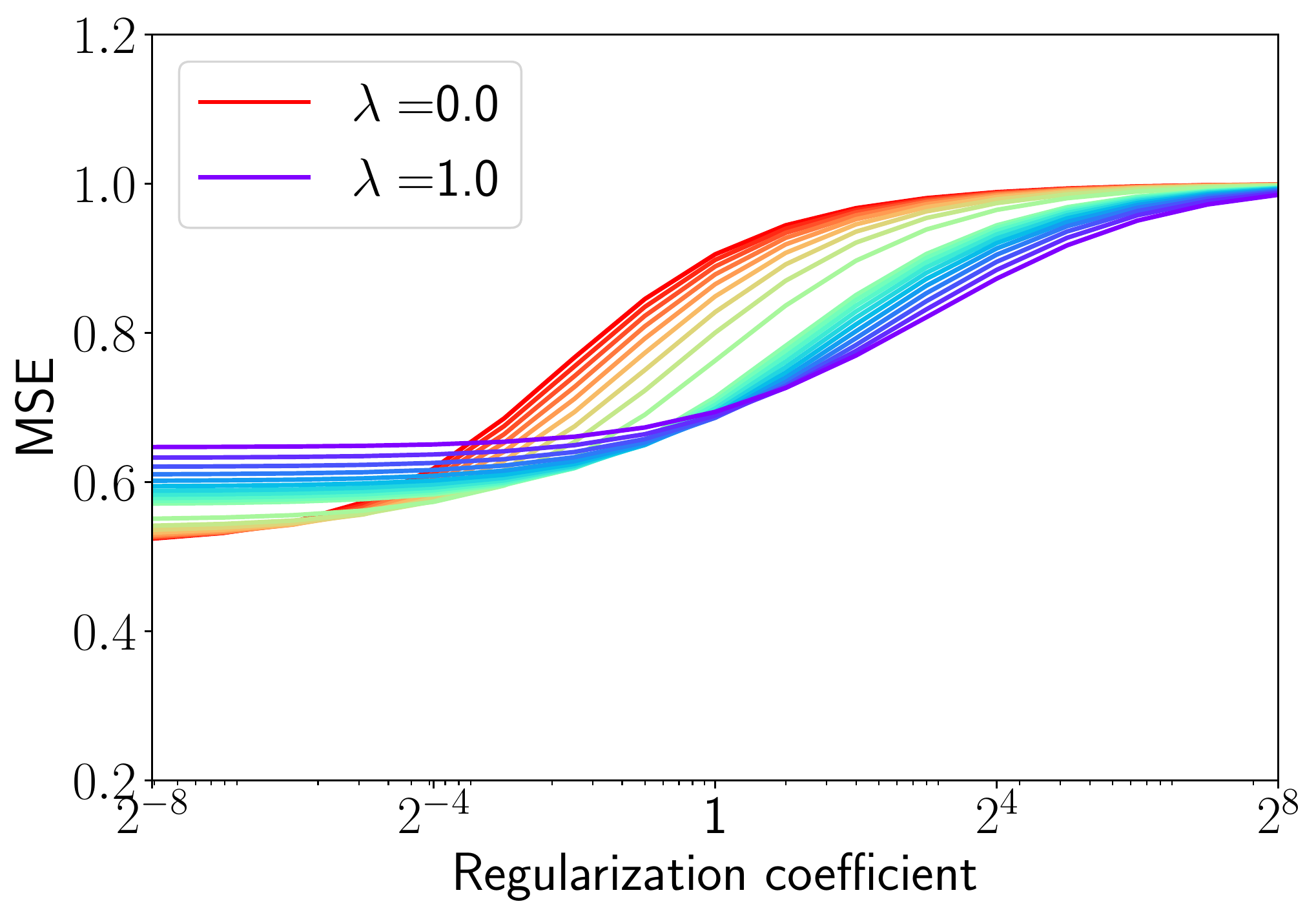}
  \end{minipage}
  \begin{minipage}{0.26\linewidth}
    \includegraphics[width=\linewidth]{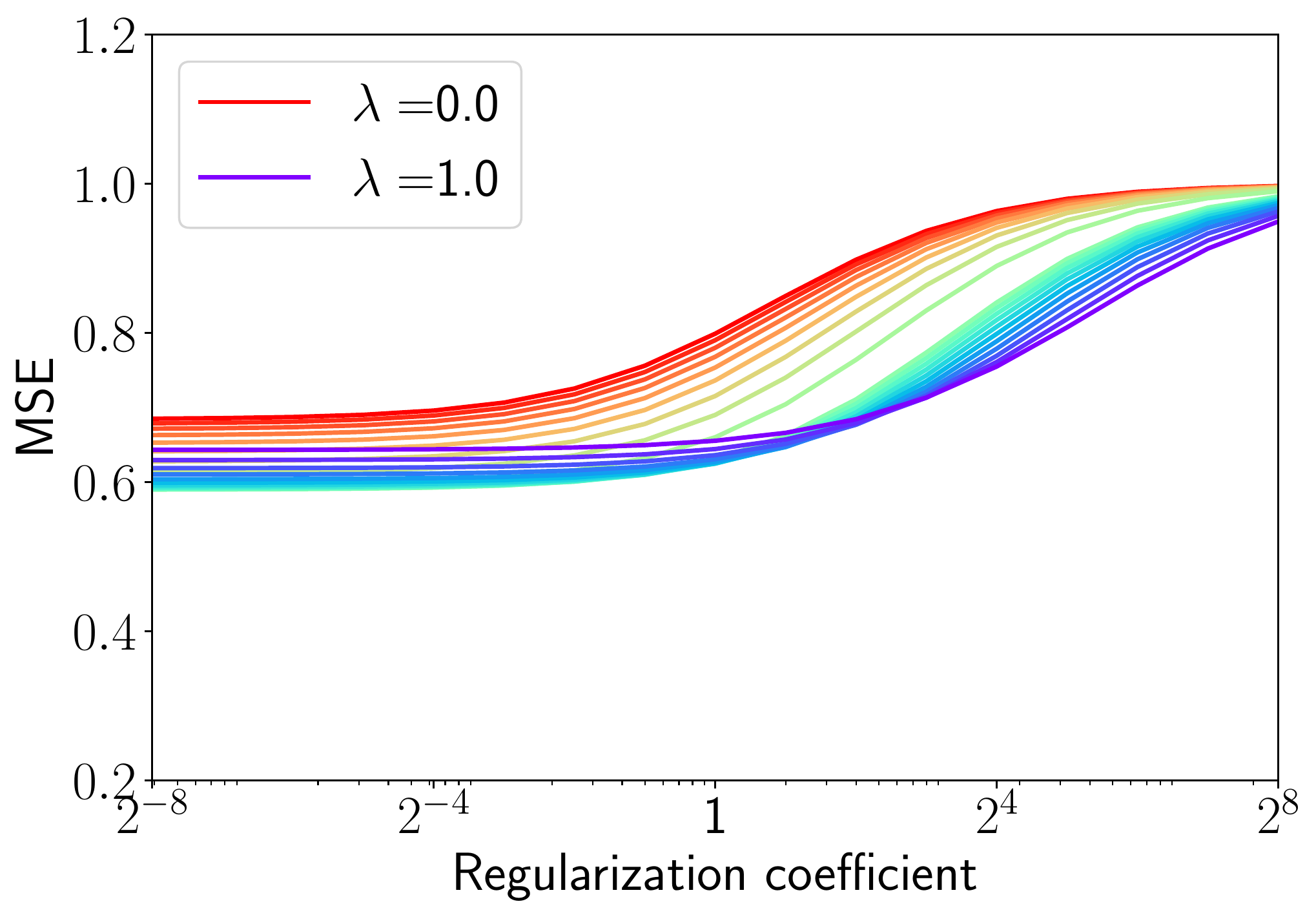}
  \end{minipage}
  \begin{minipage}{0.26\linewidth}
    \includegraphics[width=\linewidth]{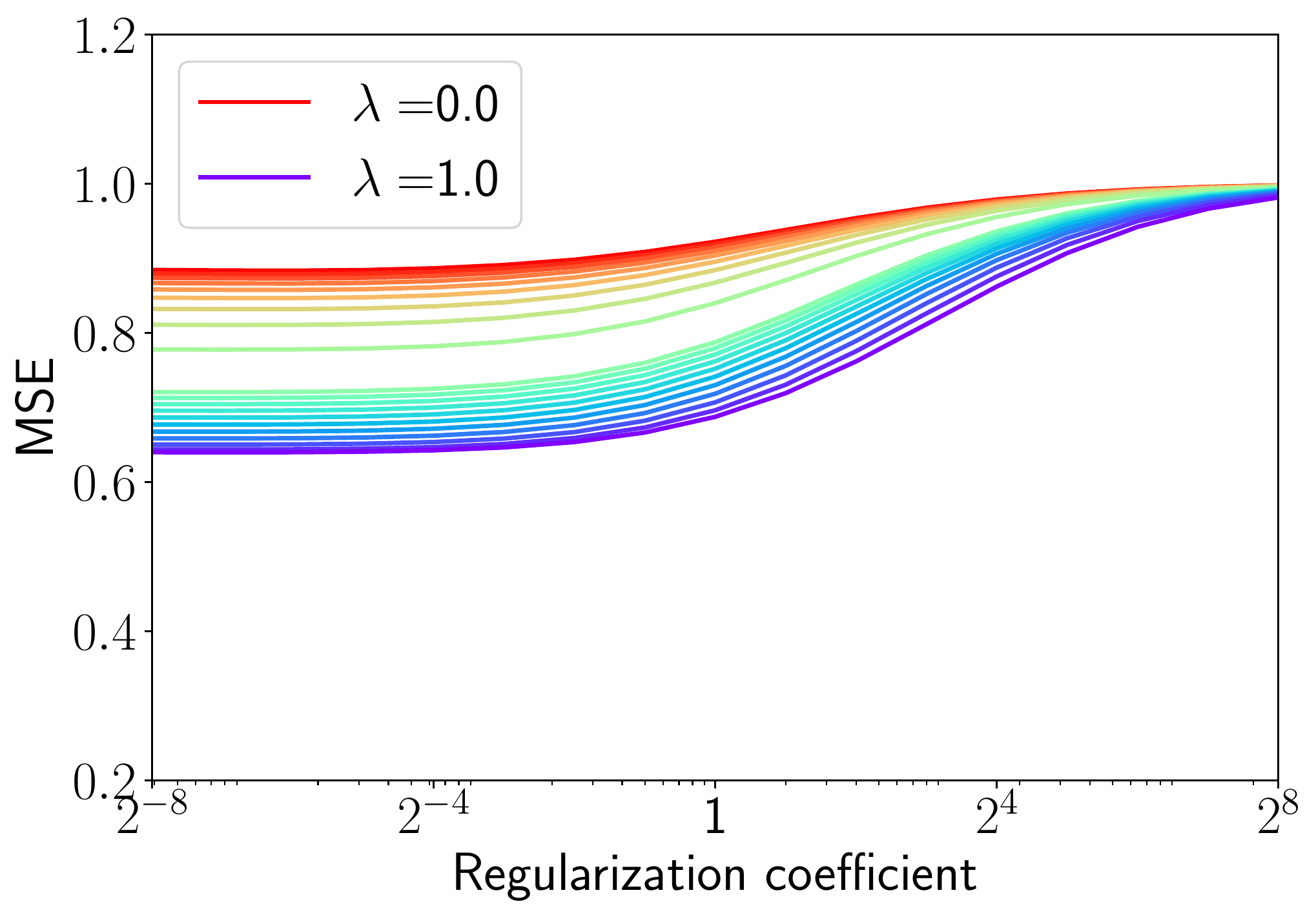}
  \end{minipage}
  \begin{minipage}{0.2\linewidth}
    \underline{Boyan's LSTD($\lambda$)}
  \end{minipage}
  \begin{minipage}{0.26\linewidth}
    \centering
    \includegraphics[width=\linewidth]{small_tabular_lstd.pdf}\\
    (a) {\em tabular} features
  \end{minipage}
  \begin{minipage}{0.26\linewidth}
    \centering
    \includegraphics[width=\linewidth]{small_binary_lstd.pdf}\\
    (b) {\em binary} features
  \end{minipage}
  \begin{minipage}{0.26\linewidth}
    \centering
    \includegraphics[width=\linewidth]{small_normal_lstd.pdf}\\
    (c) {\em non-binary} features
  \end{minipage}
  \caption{Mean squared error (MSE) of Uncorrected,
    Mixed, and Boyan's LSTD($\lambda$)
    on the {\em small} MRP $(10, 3, 0.1)$ as a function of the value
    of regularization coefficient.  Each curve shows the MSE (over 50
    runs) of a particular value of $\lambda$ for $0\le\lambda\le 1$.
    The legend only shows the color with $\lambda\in\{0,1\}$, but the
    intermediate values of $\lambda$ follow the
    color map of rainbow.  This figure is analogous to Figure 10 from
    \citet{TrueTDb}, but here we vary the value of regularization
    coefficient, while the step size is varied in \citet{TrueTDb}.}
  \label{fig:small}
\end{figure*}

\begin{figure*}[hb]
  \paragraph{Detailed results with the {\em large} MRP}
  \ \\
  \ \\
  \begin{minipage}{0.2\linewidth}
    \underline{Mixed LSTD($\lambda$)}
  \end{minipage}
  \begin{minipage}{0.26\linewidth}
    \includegraphics[width=\linewidth]{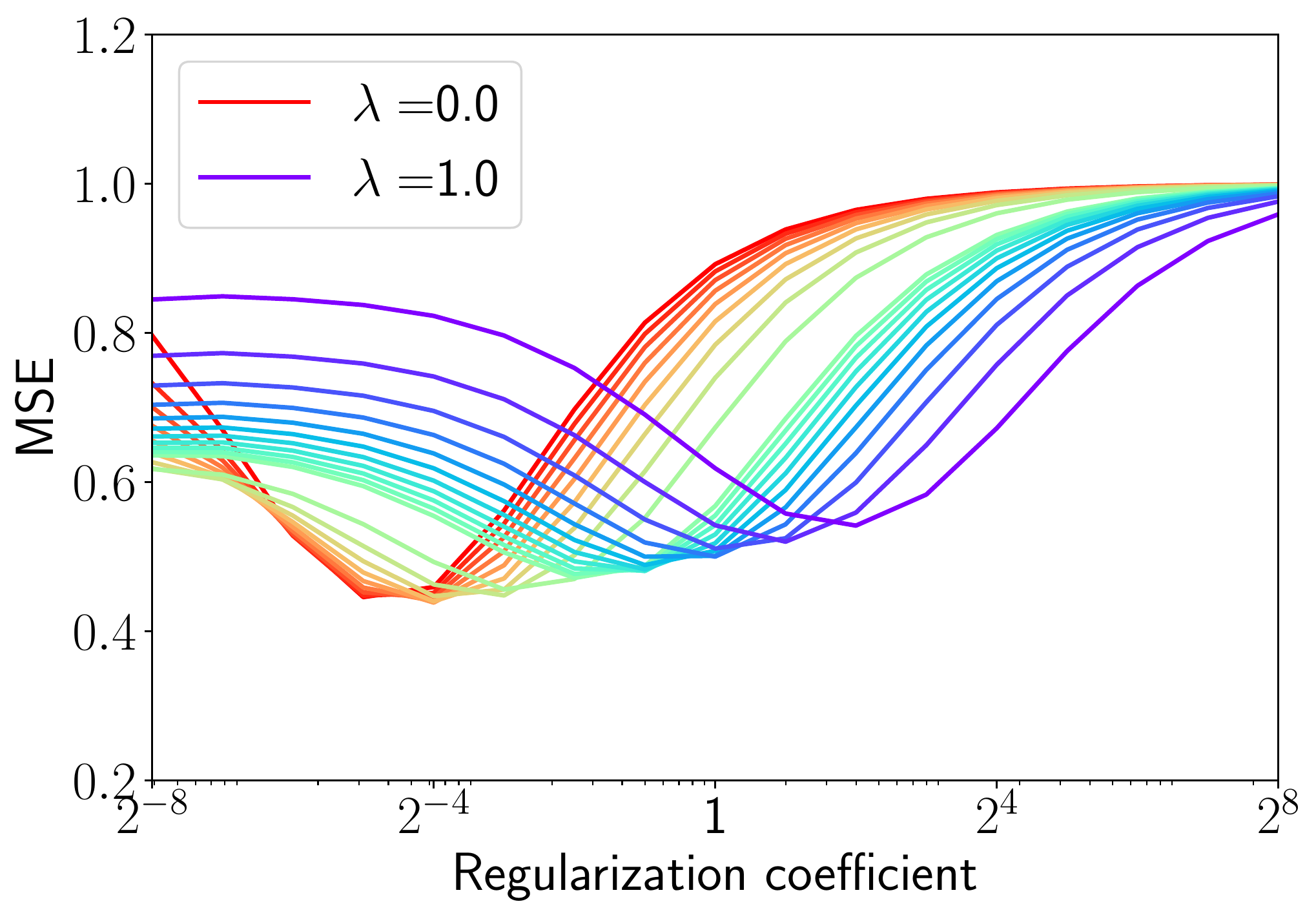}
  \end{minipage}
  \begin{minipage}{0.26\linewidth}
    \includegraphics[width=\linewidth]{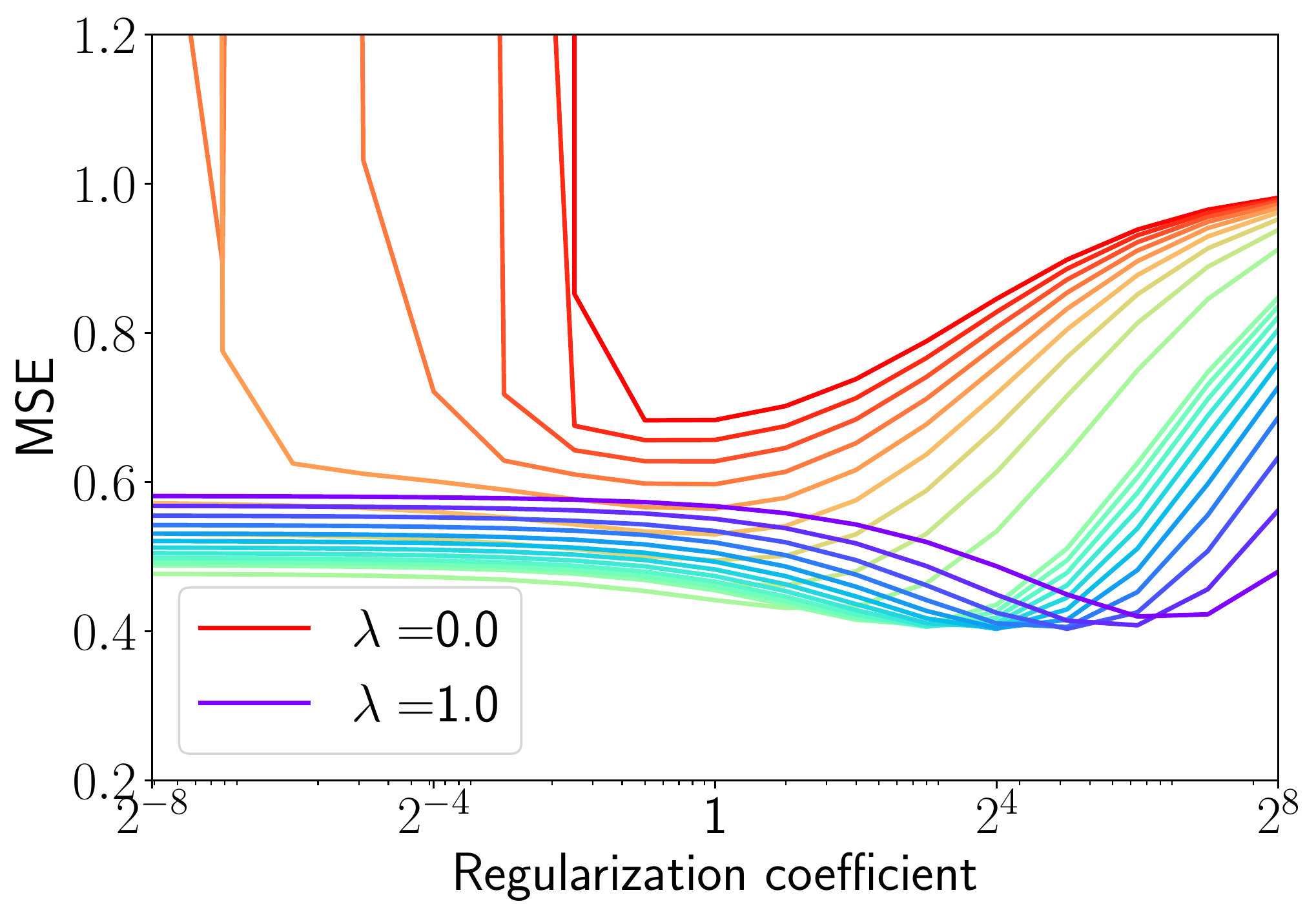}
  \end{minipage}
  \begin{minipage}{0.26\linewidth}
    \includegraphics[width=\linewidth]{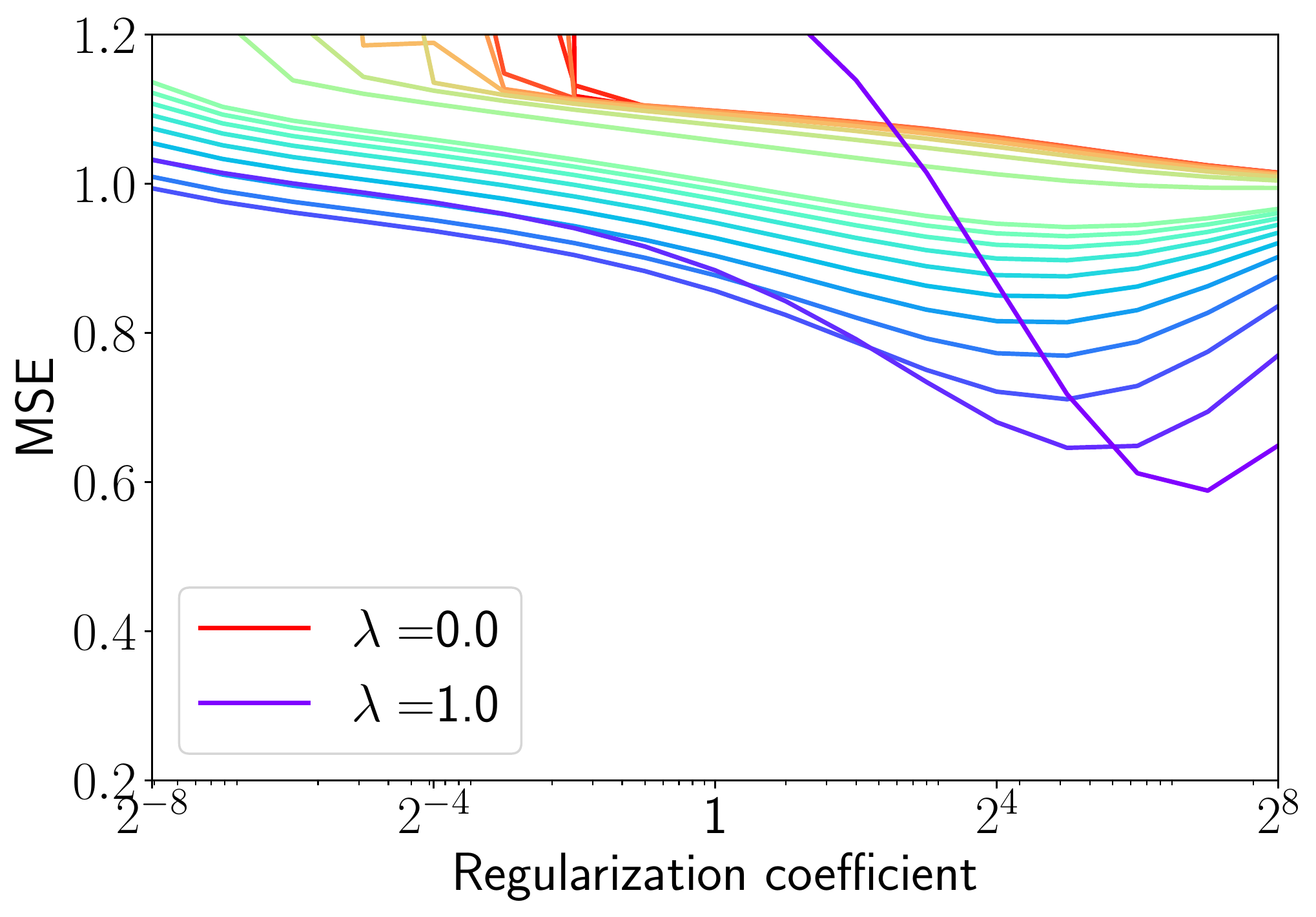}
  \end{minipage}
  \begin{minipage}{0.2\linewidth}
    \underline{Uncorrected LSTD($\lambda$)}
  \end{minipage}
  \begin{minipage}{0.26\linewidth}
    \includegraphics[width=\linewidth]{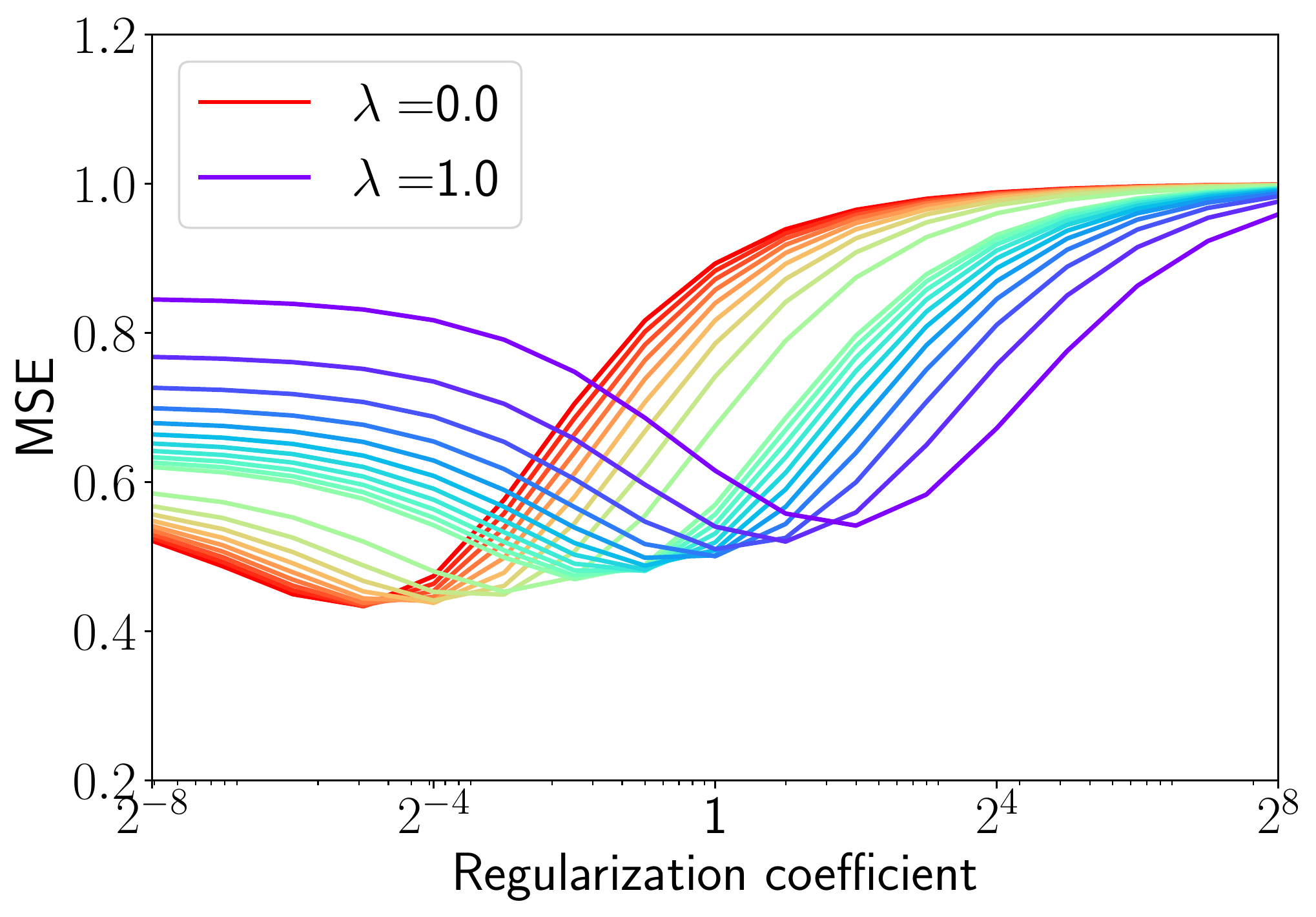}
  \end{minipage}
  \begin{minipage}{0.26\linewidth}
    \includegraphics[width=\linewidth]{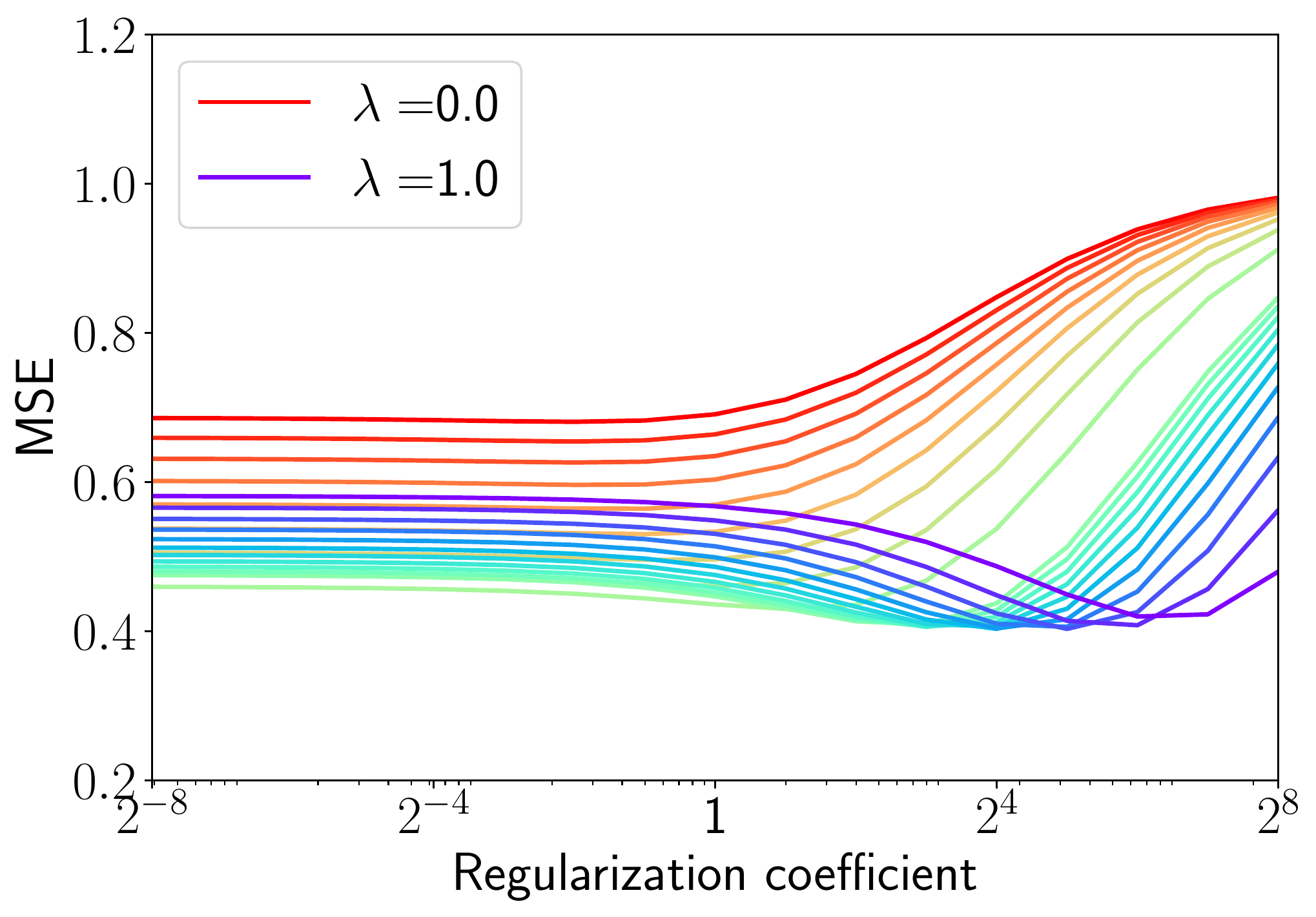}
  \end{minipage}
  \begin{minipage}{0.26\linewidth}
    \includegraphics[width=\linewidth]{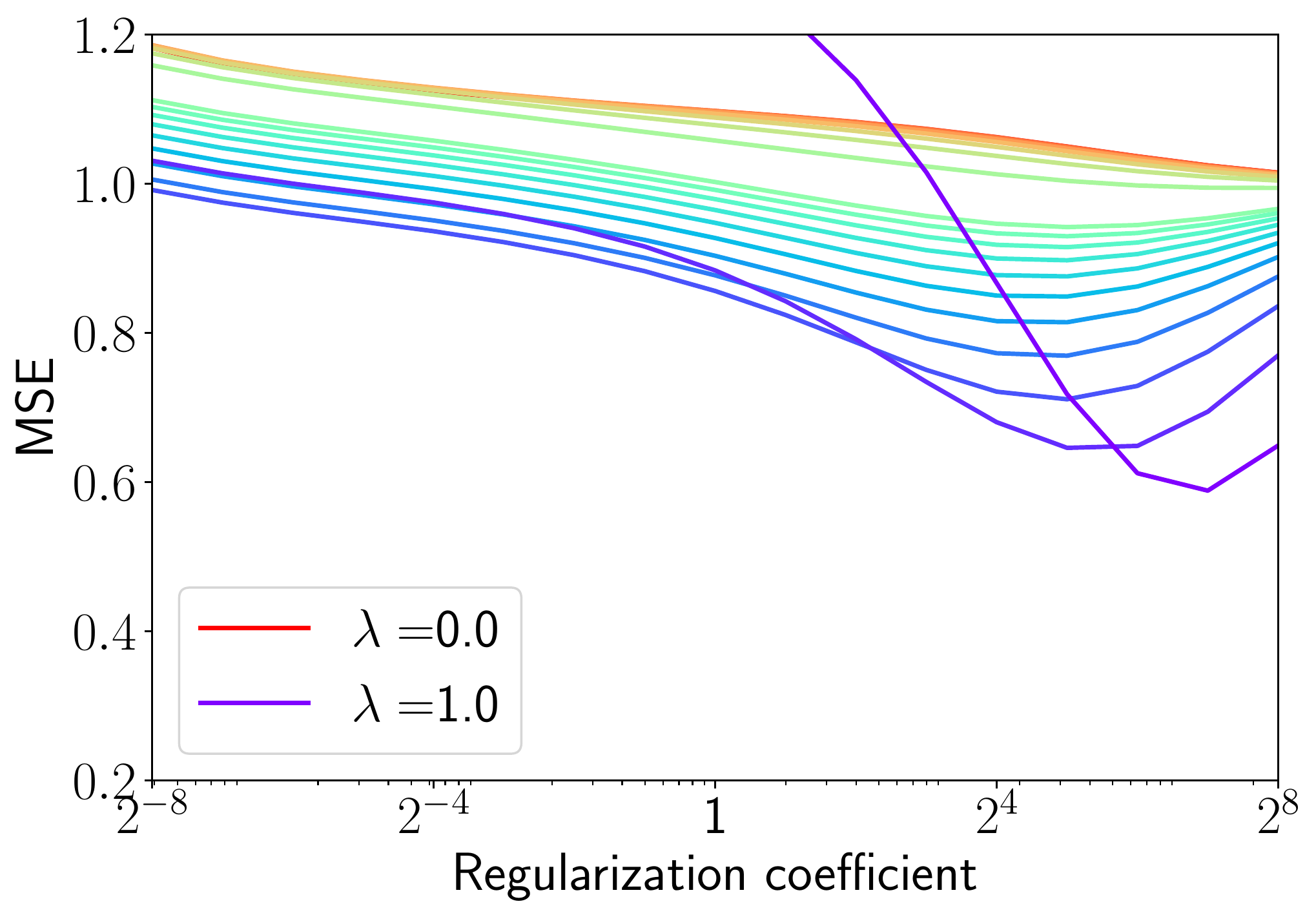}
  \end{minipage}
  \begin{minipage}{0.2\linewidth}
    \underline{Boyan's LSTD($\lambda$)}
  \end{minipage}
  \begin{minipage}{0.26\linewidth}
    \centering
    \includegraphics[width=\linewidth]{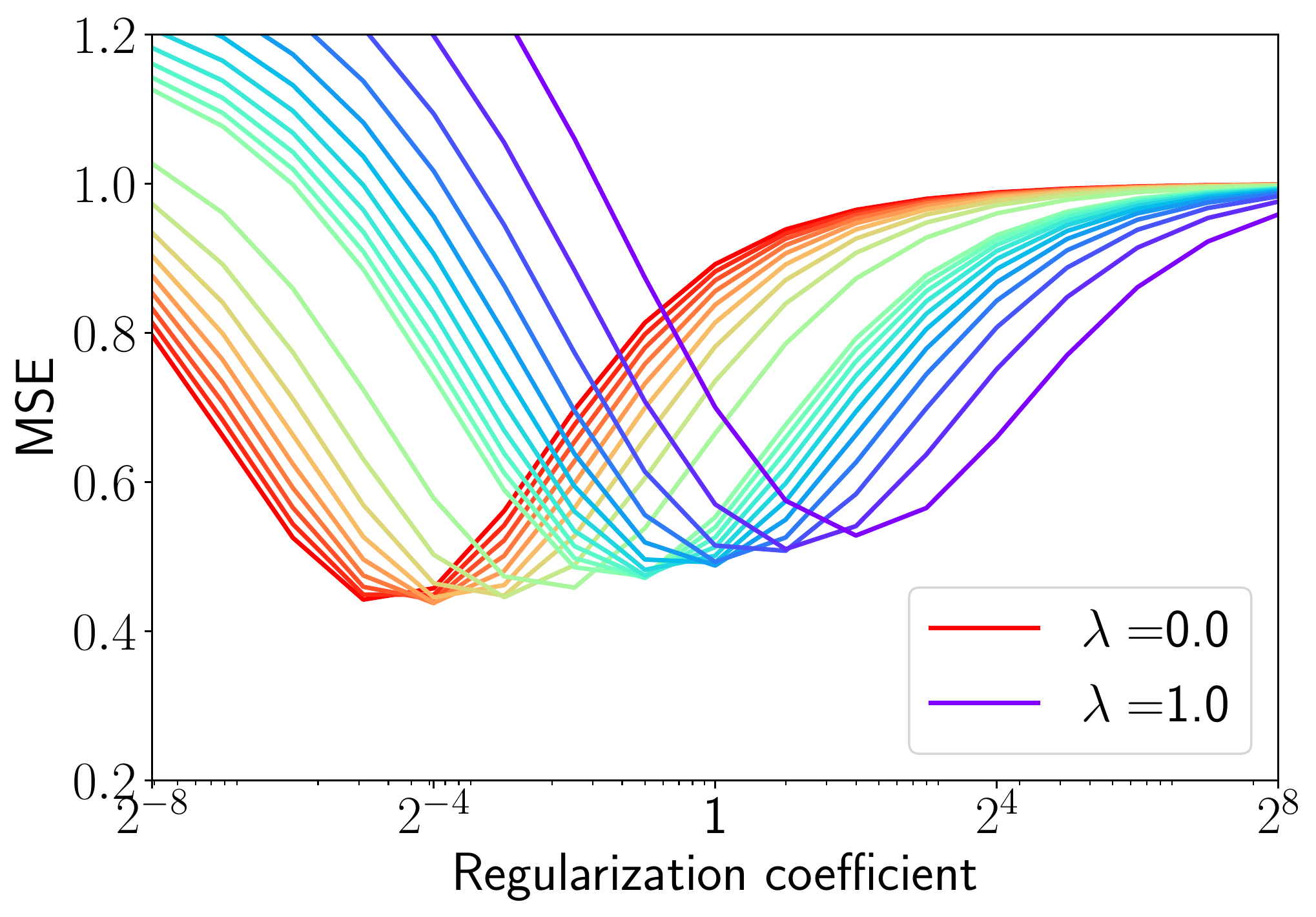}\\
    (a) {\em tabular} features
  \end{minipage}
  \begin{minipage}{0.26\linewidth}
    \centering
    \includegraphics[width=\linewidth]{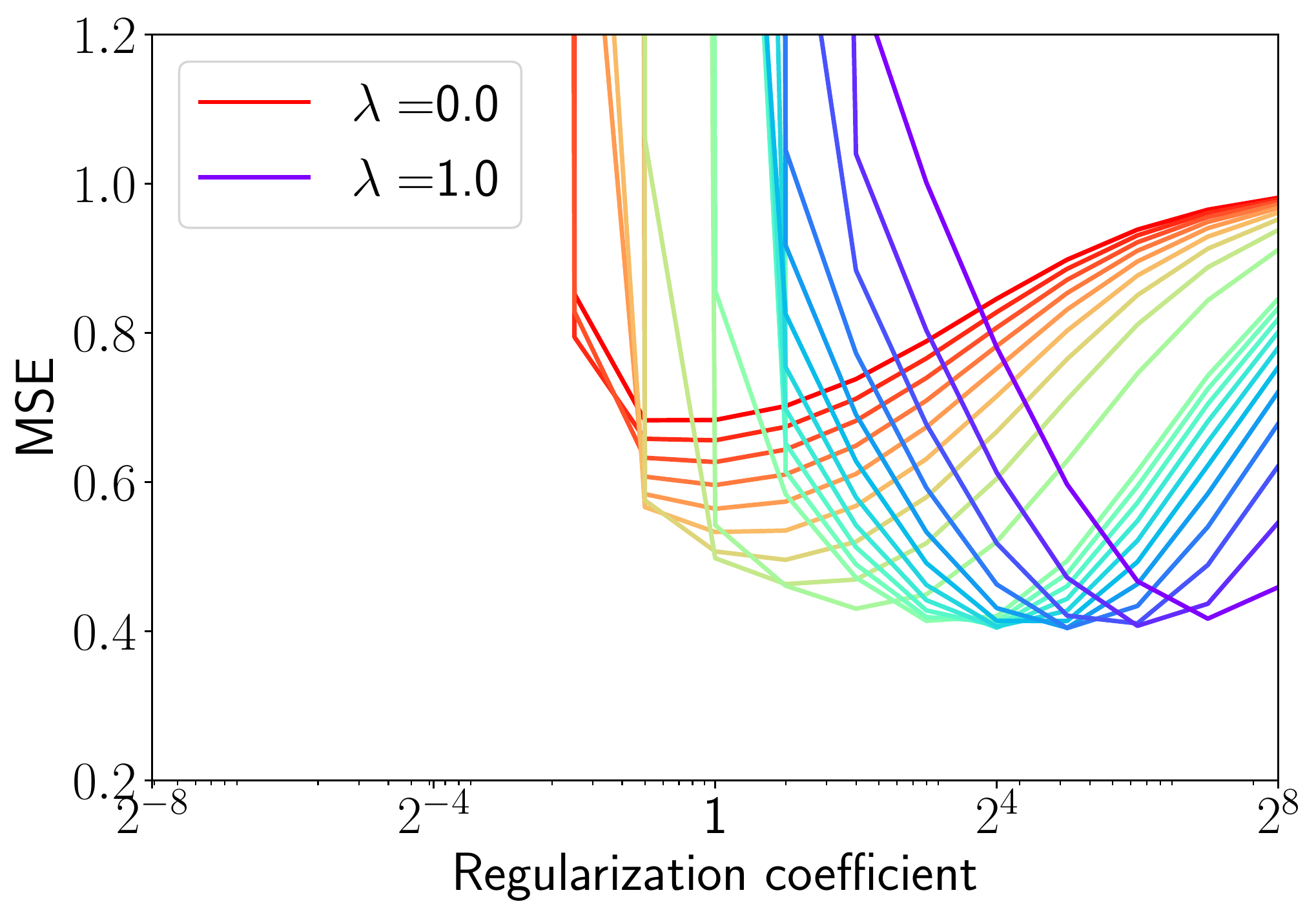}\\
    (b) {\em binary} features
  \end{minipage}
  \begin{minipage}{0.26\linewidth}
    \centering
    \includegraphics[width=\linewidth]{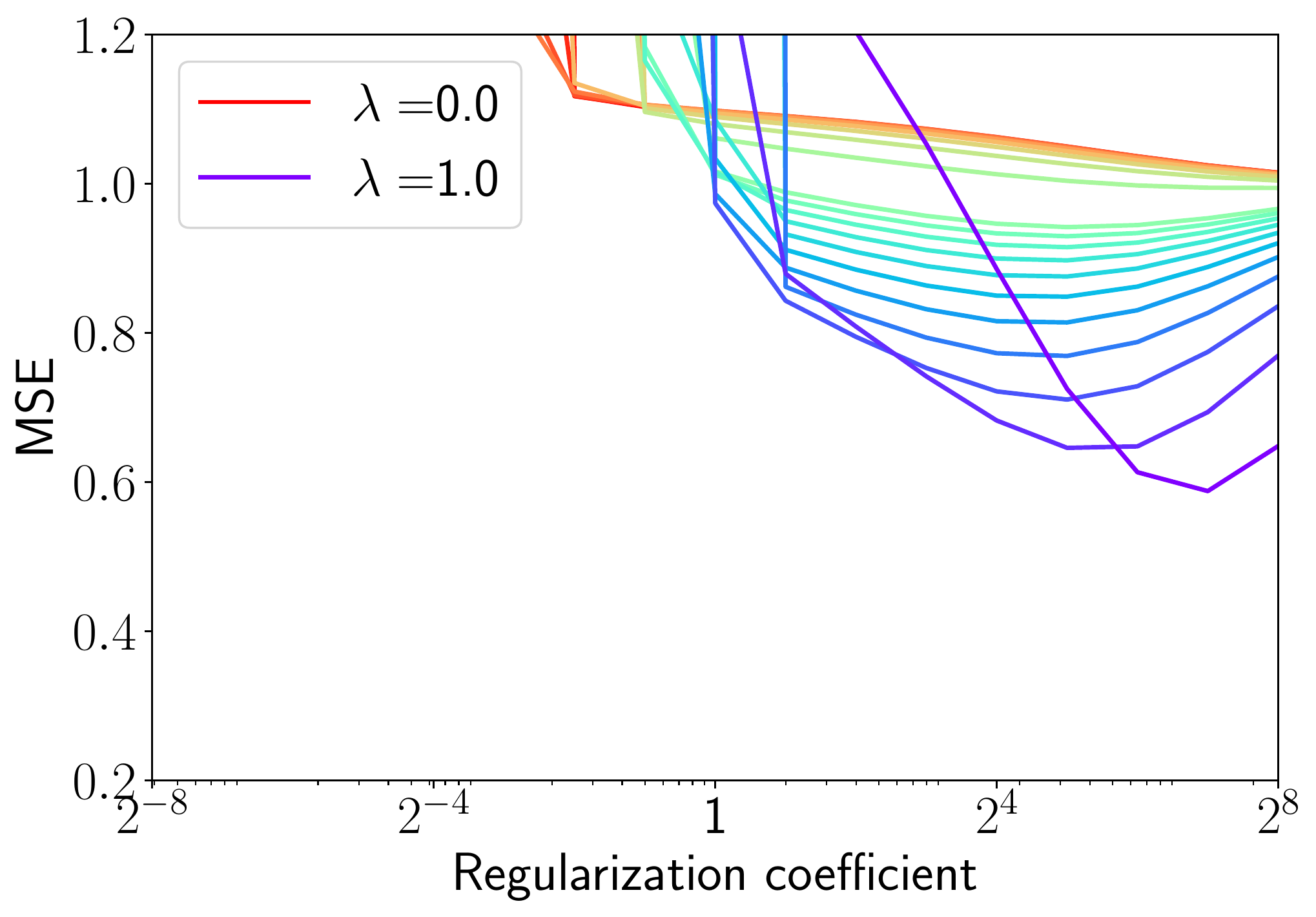}\\
    (c) {\em non-binary} features
  \end{minipage}
  \caption{Mean squared error (MSE) of Uncorrected,
    Mixed, and Boyan's LSTD($\lambda$)
    on the {\em large} MRP $(100, 10, 0.1)$ as a function of the value
    of regularization coefficient.  See the caption of \ref{fig:small}
    for the settings of experiments.}
  \label{fig:large}
\end{figure*}

\begin{figure*}[hb]
  \paragraph{Detailed results with the {\em deterministic} MRP}
  \ \\
  \ \\
  \begin{minipage}{0.2\linewidth}
    \underline{Mixed LSTD($\lambda$)}
  \end{minipage}
  \begin{minipage}{0.26\linewidth}
    \includegraphics[width=\linewidth]{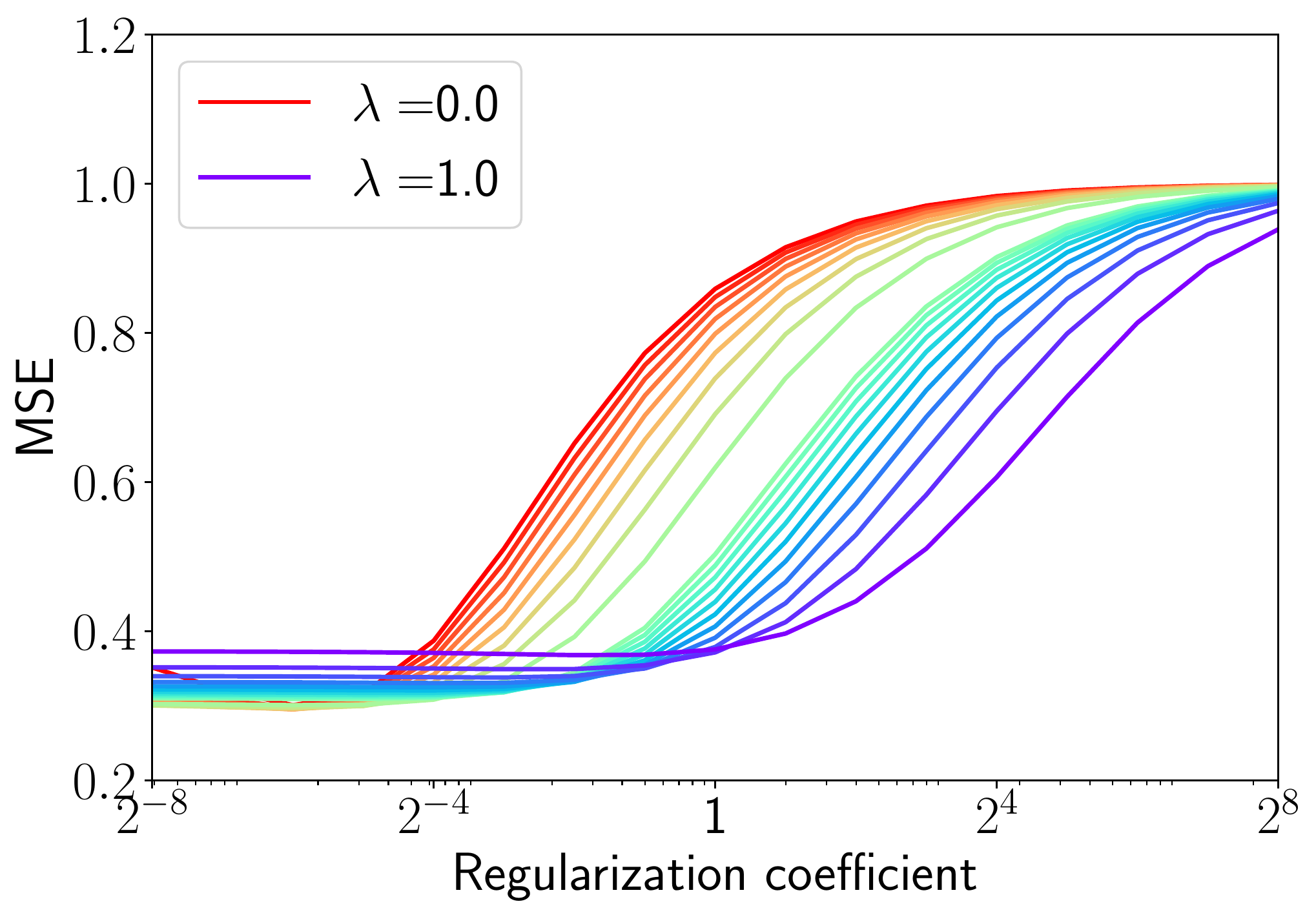}
  \end{minipage}
  \begin{minipage}{0.26\linewidth}
    \includegraphics[width=\linewidth]{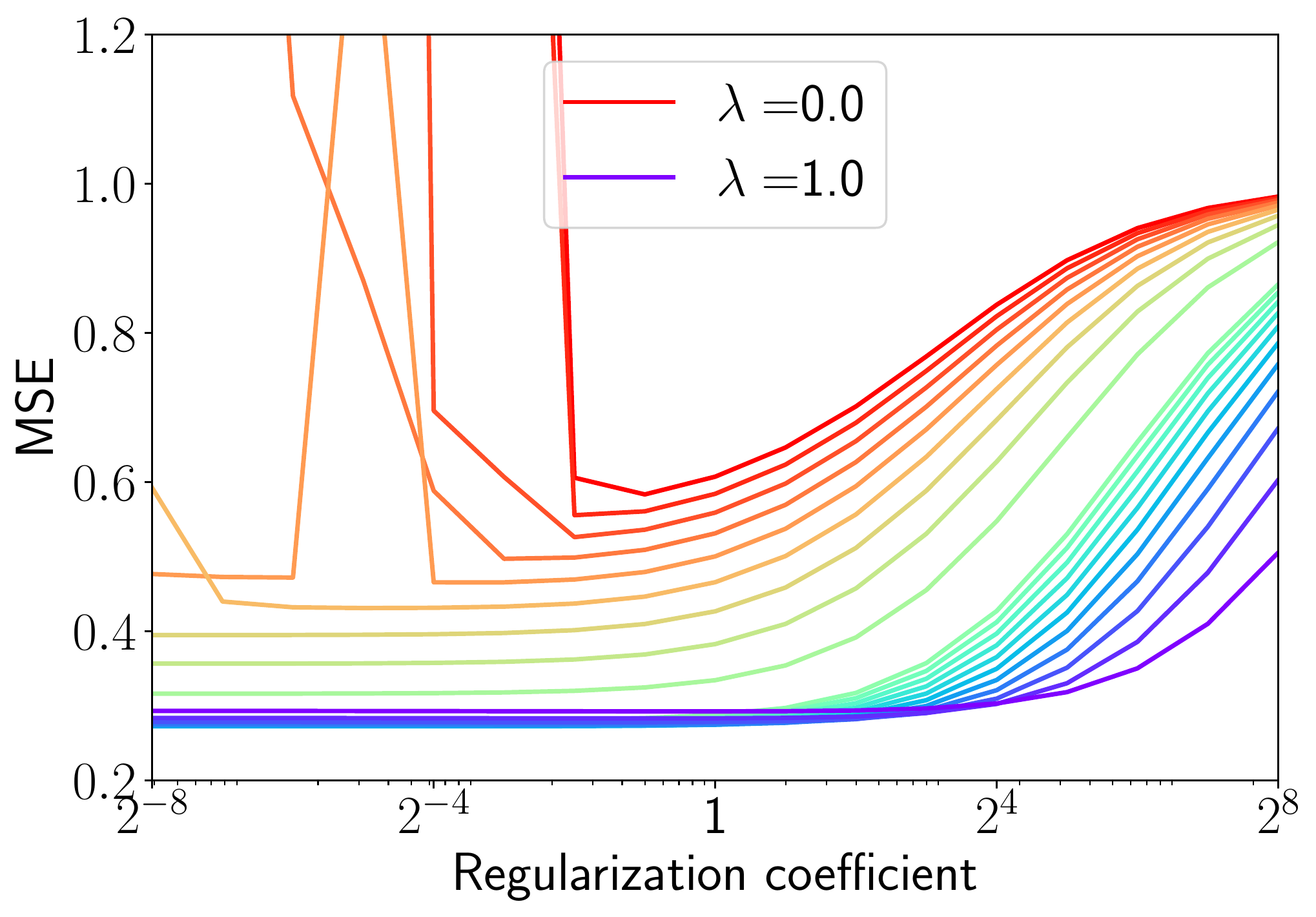}
  \end{minipage}
  \begin{minipage}{0.26\linewidth}
    \includegraphics[width=\linewidth]{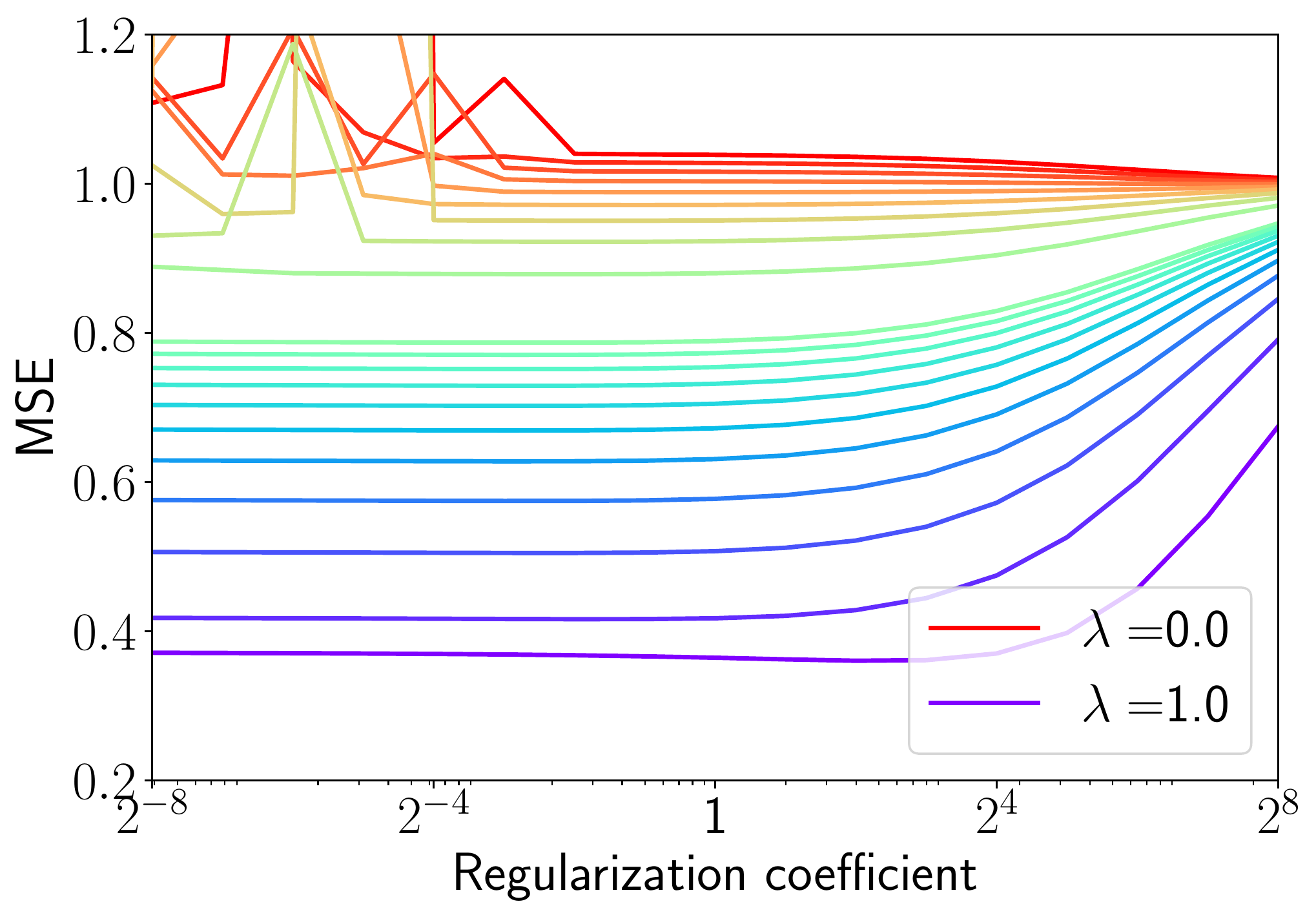}
  \end{minipage}
  \begin{minipage}{0.2\linewidth}
    \underline{Uncorrected LSTD($\lambda$)}
  \end{minipage}
  \begin{minipage}{0.26\linewidth}
    \includegraphics[width=\linewidth]{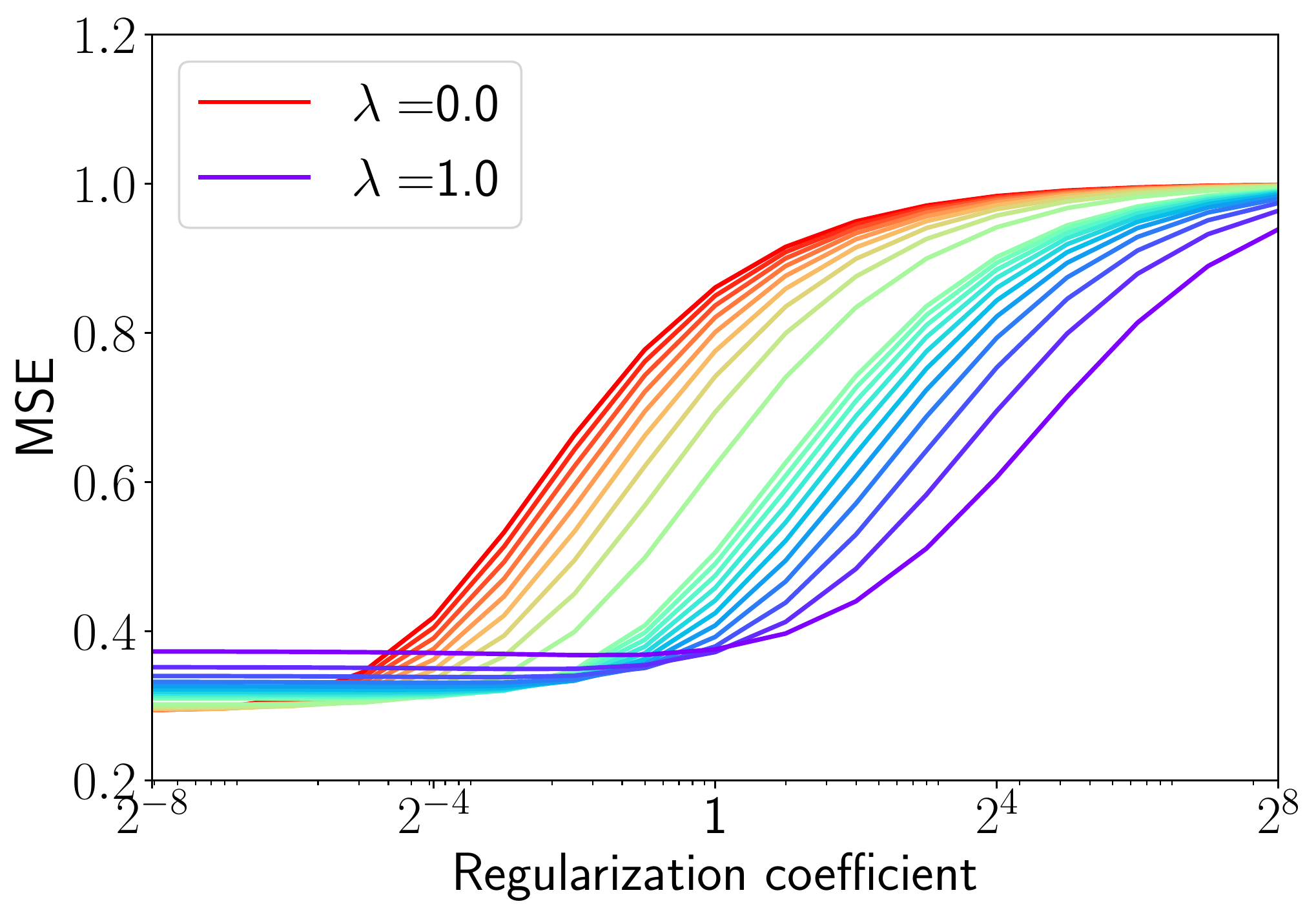}
  \end{minipage}
  \begin{minipage}{0.26\linewidth}
    \includegraphics[width=\linewidth]{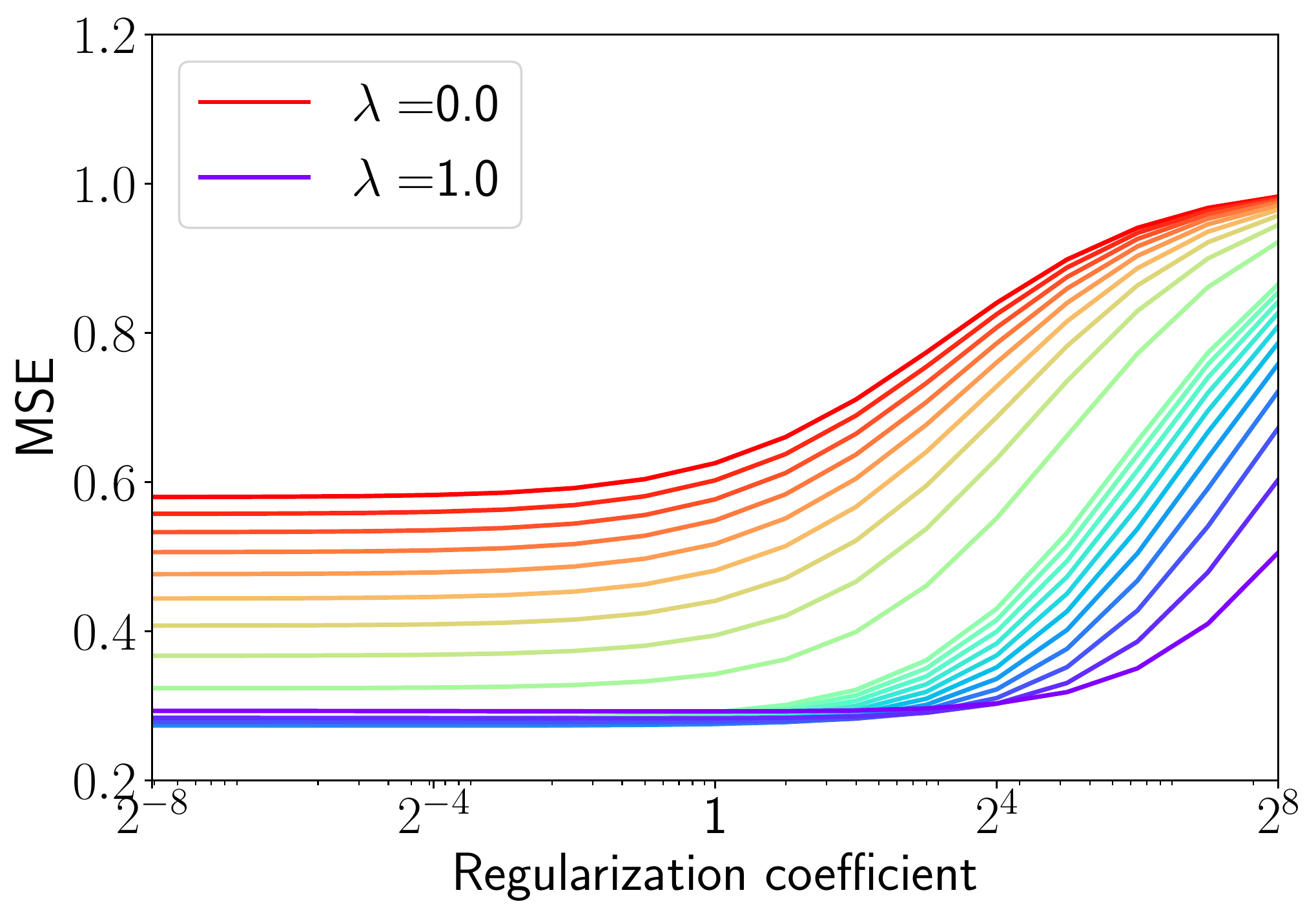}
  \end{minipage}
  \begin{minipage}{0.26\linewidth}
    \includegraphics[width=\linewidth]{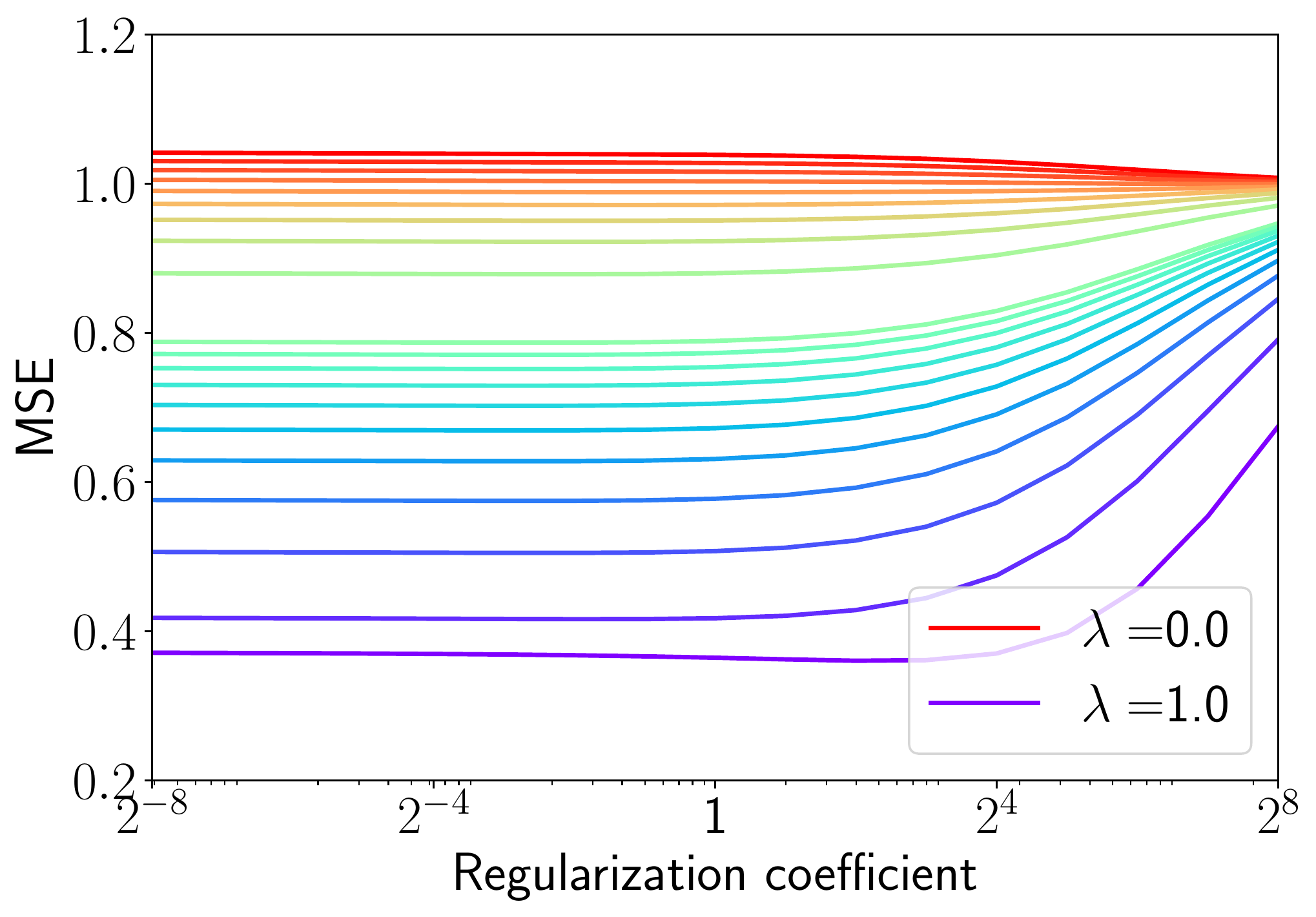}
  \end{minipage}
  \begin{minipage}{0.2\linewidth}
    \underline{Boyan's LSTD($\lambda$)}
  \end{minipage}
  \begin{minipage}{0.26\linewidth}
    \centering
    \includegraphics[width=\linewidth]{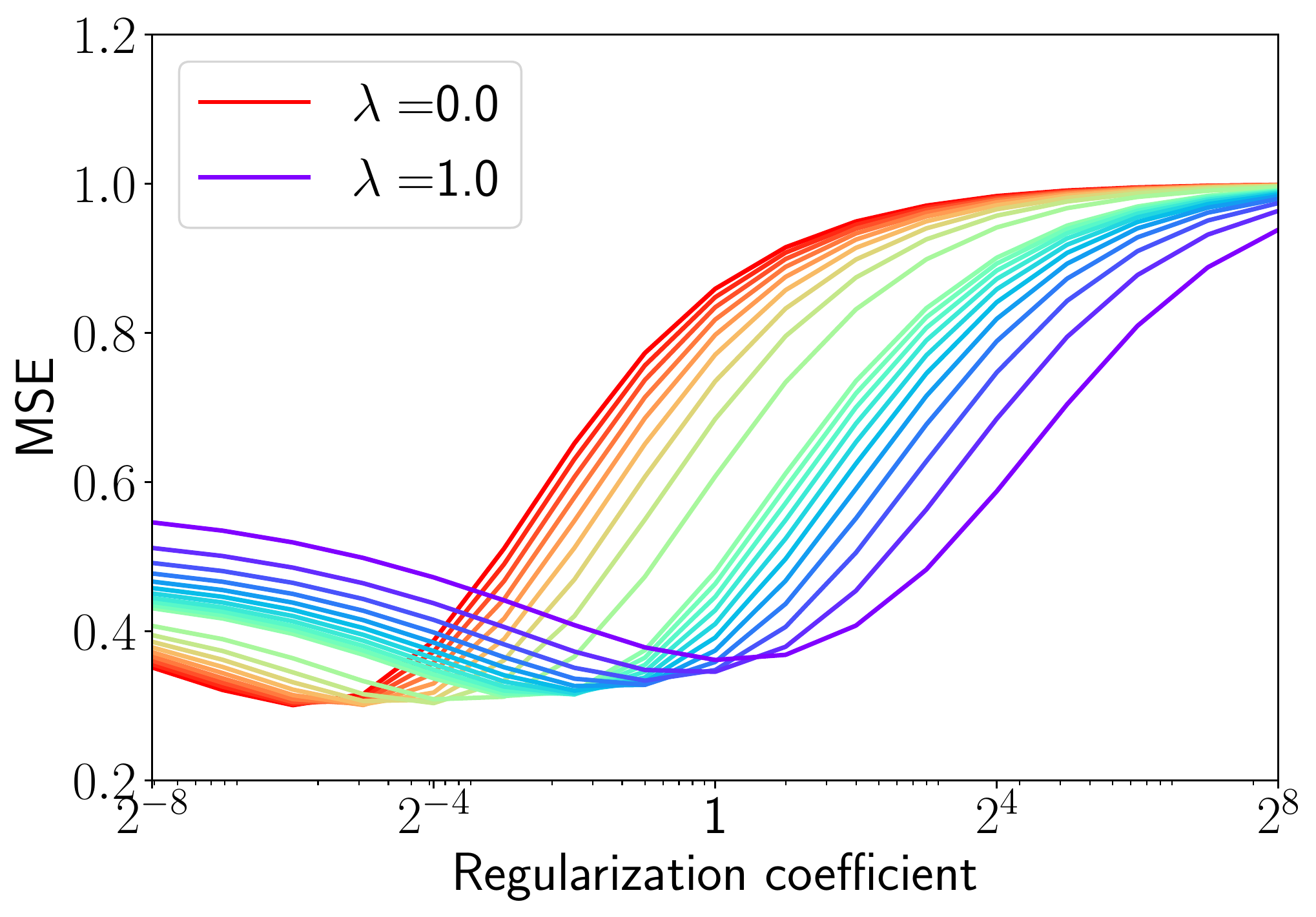}\\
    (a) {\em tabular} features
  \end{minipage}
  \begin{minipage}{0.26\linewidth}
    \centering
    \includegraphics[width=\linewidth]{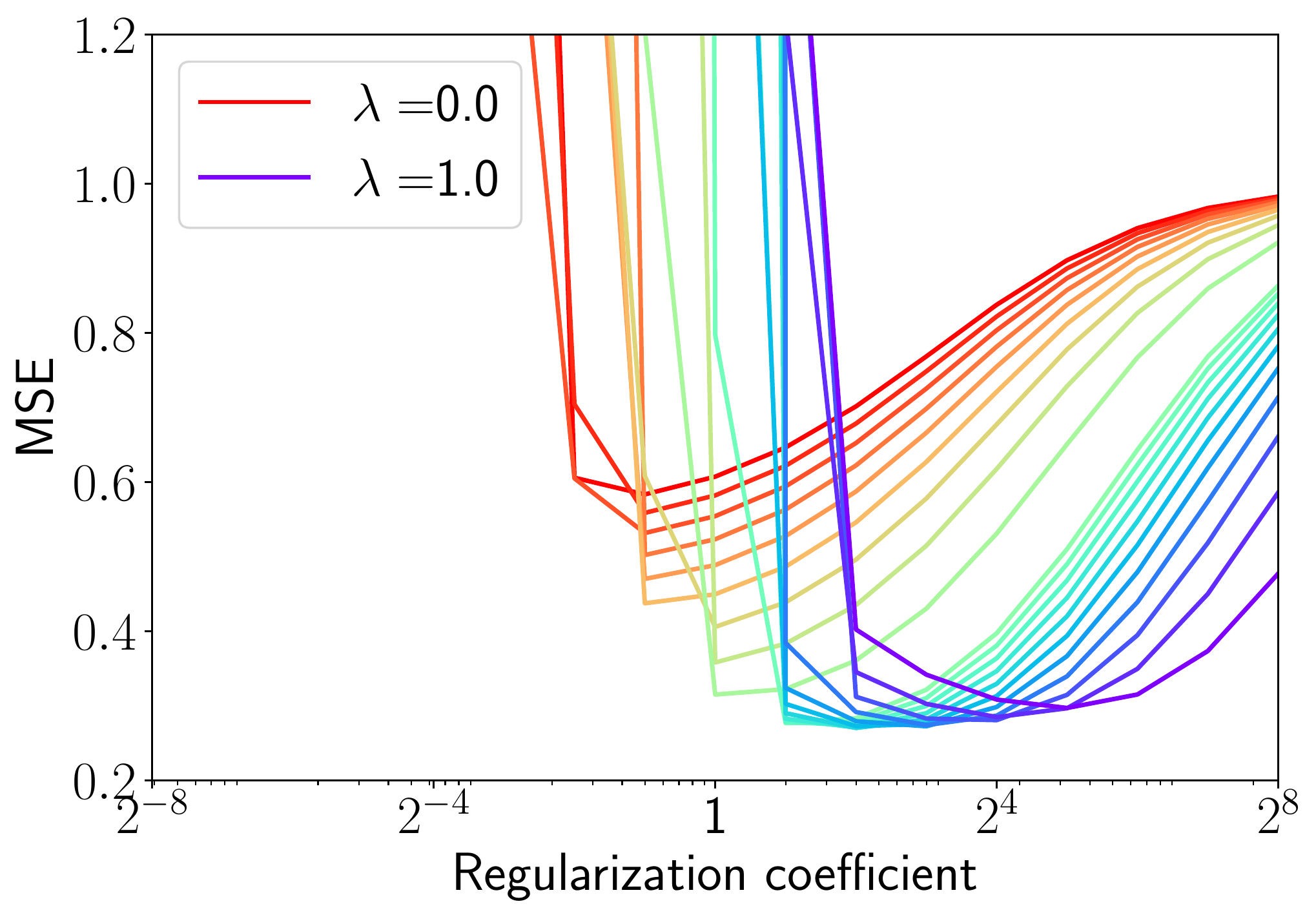}\\
    (b) {\em binary} features
  \end{minipage}
  \begin{minipage}{0.26\linewidth}
    \centering
    \includegraphics[width=\linewidth]{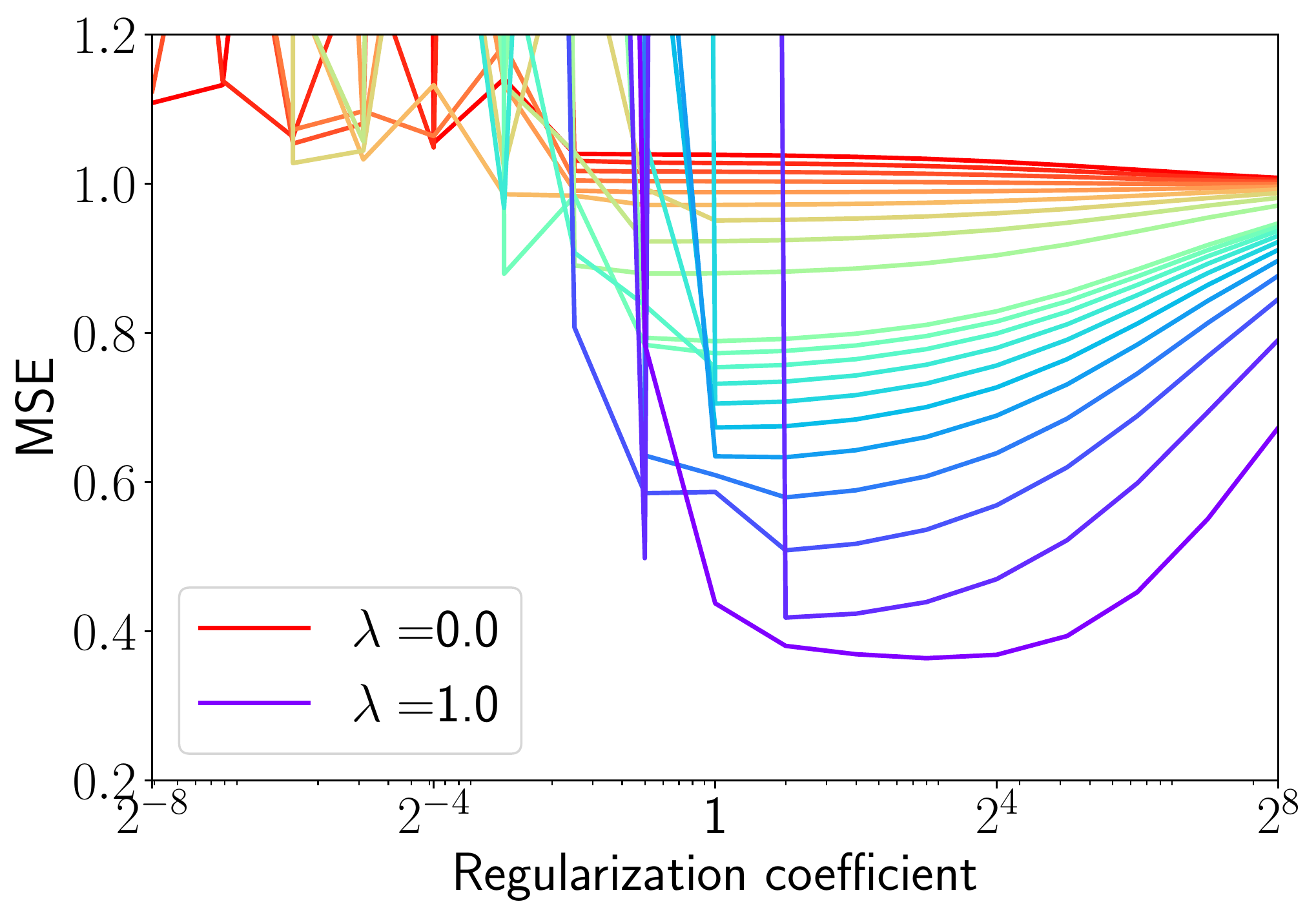}\\
    (c) {\em non-binary} features
  \end{minipage}
  \caption{Mean squared error (MSE) of Uncorrected,
    Mixed, and Boyan's LSTD($\lambda$)
    on the {\em deterministic} MRP $(100, 3, 0)$ as a function of the
    value of regularization coefficient.  See the caption of
    \ref{fig:small} for the settings of experiments.}
  \label{fig:deterministic}
\end{figure*}

Figure~\ref{fig:errorbar} thus needs to be understood with the
sensitivity of the performance to the particular values of
hyperparameters, which are shown in
Figures~\ref{fig:small}-\ref{fig:deterministic}.  These figures may be
compared against Figures~10-12 of \citet{TrueTDb}.  Note, however, that
the horizontal axis is regularization coefficient in
Figures~\ref{fig:small}-\ref{fig:deterministic} but step size in
Figures~10-12 of \citet{TrueTDb}.

Table~\ref{tbl:small} shows the computational time (in seconds) of
each LSTD($\lambda$) in each run of the experiment with the {\em
  small} MRP shown with Figure~\ref{fig:small}.
Likewise, Tables~\ref{tbl:large}-\ref{tbl:deterministic}
show the computational time with the {\em large} and {\em deterministic}
MRPs shown with Figures~\ref{fig:large}-\ref{fig:deterministic}.
Notice that each run
consists of $17\times 20=340$ combinations of the values of $\alpha$
and $\lambda$ (recall that we vary $\alpha$ in $\{2^i \mid
i=-8,-7,\ldots,7,8\}$ for each of $\lambda$ in $\{i/100\mid
i=0,10,\ldots,90,91,\ldots,100\}$).  As a reference, we also include
the running time of true online TD($\lambda$) on our environment.
Because true online TD($\lambda$) considers $30\times 20=600$
combinations of the values of hyperparameters as in \citet{TrueTDb},
the running time is normalized by $340/600$ after running all of the
combinations for fair comparison.

In our implementation, Uncorrected LSTD($\lambda$) requires slightly more
computational time than Boyan's, because at each step
Uncorrected LSTD($\lambda$) copies and stores the eligibility trace to be used
in the next step.  As is expected, Mixed LSTD($\lambda$) requires
(20~\% to 65~\%) more computational time than the other
LSTD($\lambda$)s, because Mixed LSTD($\lambda$) applies the rank-one
update twice.

\clearpage

\begin{table*}[hb]
  \centering
  \begin{tabular}{@{}rrrrr@{}}
    \toprule
    & Boyan's & Uncorrected & Mixed & true online TD($\lambda$) \\
    \midrule
    {\em tabular}    & $1.10\pm0.01$ & $1.24\pm0.01$ & $1.49\pm0.01$ & $1.21\pm0.01$ \\
    {\em binary}     & $1.54\pm0.01$ & $1.72\pm0.03$ & $2.11\pm0.03$ & $1.87\pm0.03$ \\
    {\em non-binary} & $1.54\pm0.03$ & $1.73\pm0.01$ & $2.13\pm0.02$ & $1.88\pm0.01$ \\
    \bottomrule\\
  \end{tabular}
  \caption{The average computational time (seconds) for each run,
    consisting of 340 combinations of the values of hyperparameters,
    in the experiments with the {\em small} MRP.  Here, the
    computational time of true online TD($\lambda$) is normalized by
    $340/600$ after running all of the 600 combinations of the values
    of hyperparameters.}
  \label{tbl:small}
\end{table*}

\clearpage

\begin{table*}[hb]
  \centering
  \begin{tabular}{@{}rrrrr@{}}
    \toprule
    & Boyan's & Uncorrected & Mixed & true online TD($\lambda$) \\
    \midrule
    {\em tabular}    & $27.1\pm0.6$ & $29.5\pm0.4$ & $44.7\pm1.0$ & $21.3\pm0.2$ \\
    {\em binary}     & $15.8\pm0.1$ & $17.7\pm0.1$ & $22.1\pm0.4$ & $19.8\pm0.2$ \\
    {\em non-binary} & $15.7\pm0.3$ & $17.3\pm0.1$ & $21.9\pm0.3$ & $19.8\pm0.3$ \\
    \bottomrule\\
  \end{tabular}
  \caption{The average computational time (seconds) for each run
    in the experiments with the {\em large} MRP.  See the
    caption for Table~\ref{tbl:small} for details.}
  \label{tbl:large}
\end{table*}

\clearpage

\begin{table*}[hb]
  \centering
  \begin{tabular}{@{}rrrrr@{}}
    \toprule
    & Boyan's & Uncorrected & Mixed & true online TD($\lambda$) \\
    \midrule
    {\em tabular}    & $26.6\pm0.3$ & $28.6\pm0.3$ & $43.6\pm0.6$ & $19.9\pm0.2$ \\
    {\em binary}     & $14.8\pm0.2$ & $16.2\pm0.1$ & $21.4\pm0.3$ & $18.4\pm0.4$ \\
    {\em non-binary} & $14.5\pm0.2$ & $16.3\pm0.1$ & $21.0\pm0.2$ & $18.7\pm0.2$ \\
    \bottomrule\\
  \end{tabular}
  \caption{The average computational time (seconds) for each run
    in the experiments with the {\em deterministic} MRP.  See the
    caption for Table~\ref{tbl:small} for details.}
  \label{tbl:deterministic}
\end{table*}

\bibliography{lstd2}
\bibliographystyle{abbrvnat}

\end{document}